\documentclass[10pt,twocolumn]{article} 
\usepackage{simpleConference}
\usepackage{times}
\usepackage{graphicx}
\usepackage{amssymb}
\usepackage{url,hyperref}
\hypersetup{
    linkcolor=black,
    hidelinks
    }
\usepackage{booktabs}
\usepackage{multirow}

\graphicspath{{media/}}     

\begin{document}

\title{Machine Learning Approaches for Diagnostics and Prognostics of Industrial Systems Using Open Source Data from PHM Data Challenges: A Review}

\author{Hanqi Su\thanks{\textit{Corresponding author}}, Jay Lee \\
Center for Industrial Artificial Intelligence, Department of Mechanical Engineering \\
University of Maryland, College Park, MD, USA \\
\texttt{\{hanqisu,leejay\}@umd.edu}  \\
}

\maketitle
\thispagestyle{empty}

\begin{abstract}
In the field of Prognostics and Health Management (PHM), recent years have witnessed a significant surge in the application of machine learning (ML). Despite this growth, the field grapples with a lack of unified guidelines and systematic approaches for effectively implementing these ML techniques and comprehensive analysis regarding industrial open-source data across varied scenarios. To address these gaps, this paper provides a comprehensive review of ML approaches for diagnostics and prognostics of industrial systems using open-source datasets from PHM Data Challenge Competitions held between 2018 and 2023 by PHM Society and IEEE Reliability Society and summarizes a unified ML framework. This review systematically categorizes and scrutinizes the problems, challenges, methodologies, and advancements demonstrated in these competitions, highlighting the evolving role of both conventional machine learning and deep learning in tackling complex industrial tasks related to detection, diagnosis, assessment, and prognosis. Moreover, this paper delves into the common challenges in PHM data challenge competitions by emphasizing data-related and model-related issues and evaluating the limitations of these competitions. The potential solutions to address these challenges are also summarized. Finally, we identify key themes and potential directions for future research, providing opportunities and prospects for next-generation ML-PHM development in PHM domain.
\end{abstract}

\section{Introduction}
In the era of Industry 4.0, the emphasis on the reliability, efficiency, and longevity of industrial systems has become crucial~\cite{lasi2014industry}. Prognostics and Health Management (PHM) integrates the detection, diagnosis, assessment, and prognosis of system failures to address the growing need for proactive system health management~\cite{zio2022prognostics}. This integrative approach can enhance the reliability and safety of industrial systems, with minimal unplanned downtimes and reduced maintenance costs.

The rise of the Internet of Things~\cite{sisinni2018industrial}, big data analytics~\cite{tsui2019big}, cyber-physical systems~\cite{lee2015cyber}, machine learning (ML)~\cite{huang2017review}, deep learning (DL) ~\cite{rezaeianjouybari2020deep}, and industrial artificial intelligence~\cite{peres2020industrial,lee2020industrial} has paved the way for a transformative shift in PHM. Historically, PHM methods largely relied on physical-based methods, which derived their strength from profound insights into system physics, material properties, and failure mechanisms. These approaches, however, often struggled with scalability, adaptability, and the ability to handle the vast variability and uncertainties inherent in real-world operations. In contrast, ML techniques present the capability to model complex systems comprehensively, uncover intricate patterns, and accurately diagnose and predict failures. Yet, employing ML in PHM poses several challenges, including the critical need for usable, useful, high-quality, and extensive data sets, extensive computing resources, a solid infrastructure for data collection and processing, as well as the necessity for assembly of a team of professionals proficient in artificial intelligence (AI) and ML~\cite{lee2018industrial,amershi2019software}. Despite these challenges, the momentum for implementing ML in industrial settings is evident. With the gradual move towards digital transformation within the industry, incorporating ML into PHM aligns with the principles of Industry 4.0~\cite{polverino2023machine}. It underscores the capability of data-driven approaches to provide precise and adaptable diagnostic and prognostic solutions and represents a strategic shift towards leveraging ML techniques for enhanced predictive maintenance.

\subsection{A Survey of Machine Learning Based PHM Reviews}
A multitude of comprehensive reviews on AI/ML/DL methods applied in PHM domain are noted~\cite{serradilla2022deep,qiu2023deep,kumar2023review,ochella2022artificial,yucesan2021survey,nguyen2023review,polverino2023machine,rezaeianjouybari2020deep,hazra2024prognostics}. These reviews evaluate the existing literature both qualitatively and quantitatively, presenting their distinct perspectives and pinpointing the trends and new concepts of AI/ML/DL methods for PHM across various scenarios.~\cite{serradilla2022deep} reviewed DL architectures like a one-class neural network (OCNN), self-organizing map (SOM), and generative techniques aligned with the industrial needs from a predictive maintenance perspective.~\cite{qiu2023deep} highlighted seven DL architectures, including emerging DL methods such as graph neural networks, transformers, and generative adversarial networks, addressing four different challenges (imbalanced data, multimodal data fusion, compound fault types, and edge device implementation). Additionally, ~\cite{kumar2023review} provided a comprehensive examination of PHM methods in the context of smart factories, spanning from traditional ML approaches to DL-based approaches. An extensive review on modeling techniques supporting PHM of industrial equipment, specifically within onshore wind energy and civil aviation sectors, was given by~\cite{yucesan2021survey}, wherein they discuss how modeling approaches are shaped by industry-specific factors (maintenance strategies, implementation aspects, and supporting technologies). Furthermore,~\cite{nguyen2023review} proposed a general guideline under AI-based PHM for selecting appropriate techniques to solve specific PHM problems.

\subsection{Motivation}

The aforementioned review papers in section 1.1 serve as the groundwork for further PHM development. Currently, a significant portion of the literature primarily examines algorithms based on their capabilities and functionalities, with numerous reviews covering various ML or DL methods like CNN, RNN, GAN, GNN, Transformer, etc. However, many researchers tend to rely on artificial datasets for algorithm testing in their studies, rather than using real, industry-specific datasets. Additionally, the focus of discussed data sets predominantly lies on mechanical components commonly used in algorithm development, such as gears and bearings. Furthermore, a considerable portion of the data sets mentioned in these studies are not publicly accessible, and among the available ones, most date back more than six years (before 2018). In PHM domain, organizations such as the PHM Society\footnote{\url{https://phmsociety.org/}} and the IEEE Reliability Society\footnote{\url{https://rs.ieee.org/}} have played pivotal roles in fostering innovation, research, and collaboration over the last fifteen years. It is worth noting that these organizations provide participants with different real open-source industrial datasets and pose real-world challenges by holding PHM data challenge competitions. These competitions seek to accelerate the development and validation of cutting-edge PHM methodologies, bridging the gap between academia and industry. Therefore, conducting in-depth analysis and review of industrial open source datasets is necessary to propel the evolution of data-centric techniques, ML, and AI within the PHM domain.

\subsection{Contributions}
This study aims to conduct a comprehensive problem-challe-
nge-solution-application-oriented review using the industrial open source data available in the last six years from PHM Data Challenge. In pursuit of the objective, 59 research papers were reviewed, encompassing both competition winning contributions and subsequent exploratory research undertaken post competition. The detailed paper selection and investigation are discussed in section 2.1. Our contributions include the following:

\begin{enumerate}

\item This study summarizes the problems and solutions presented in nine PHM data challenge competitions, elucidating the tasks, challenges, ML or DL methods, and analytical strategies employed to tackle these competitions.

\item We propose a unified ML framework for the PHM domain based on this review study, serving as a general guideline for the development of future ML models.

\item We discuss common challenges associated with industrial open source data, underscoring specific issues related to data issues (missing data, data imbalance, and domain shift), and model issues (model selection, machine learning model interpretability, model robustness and generalization) in data-driven approaches. Possible solutions are also provided. Moreover, we evaluate the limitations of these competitions and suggest future directions.

\item We identify five further research directions for the application of ML in PHM. These include: (1) a need for open-source multi-modal datasets, (2) development of multi-modal machine learning approaches, (3) further exploration in machine learning model interpretability, (4) novel transfer learning and domain adaptation techniques development for model robustness and generalization, and (5) potential utilization of large language models and industrial large knowledge models.

\end{enumerate}

\begin{figure*}[!ht]
\centering\includegraphics[width=\linewidth]{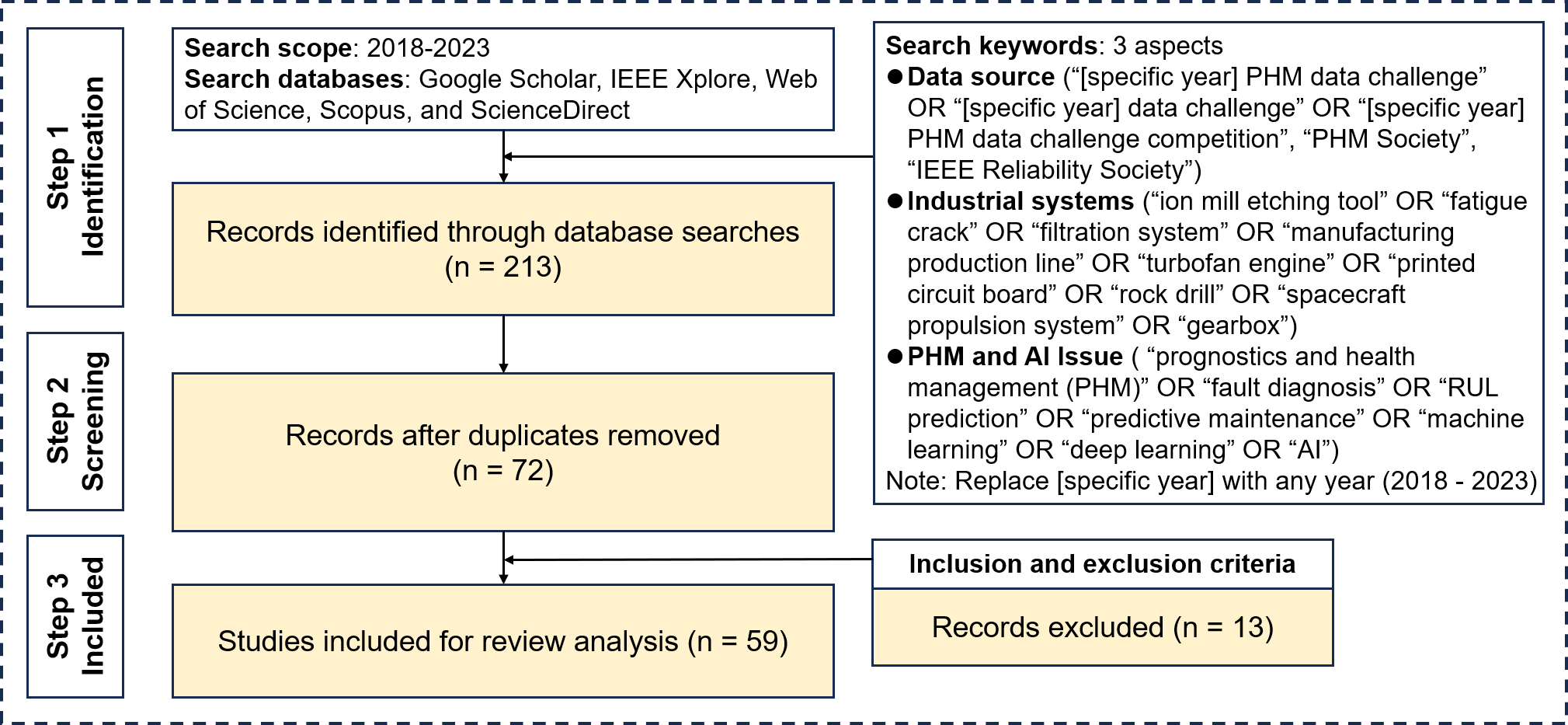}
\caption{Research Paper Selection Process Using PRISMA Method}
\label{fig:11}
\end{figure*}

The rest of this paper is organized as follows: Section 2 introduces the methodology of how we select research papers, an overview of PHM data challenge competitions, and underscores major research tasks within PHM. Section 3 respectively introduces the prevalent challenges associated with two parts: detection \& diagnosis, and assessment \& prognosis, subsequently detailing the solutions presented in various PHM data competitions individually. A unified ML framework for the PHM domain is proposed. Section 4 critically summarizes and examines common challenges in the PHM domain through the perspective of ML methodologies, addressing concerns related to data issues, and model issues. The limitations of these competitions are also discussed. Section 5 provides five research directions for future PHM development. Section 6 concludes the paper, highlighting its findings and contributions.

\section{Background}
In this section, we first introduce the procedure for selecting research papers using systematic reviews and meta-analyses (PRISMA) method. Next, we provide an overview of the PHM data challenge competition. Then, we outline the major research tasks in the PHM field and provide respective explanations. 


\subsection{Identification, Screening, and Inclusion of Studies}
For paper selection and investigation, this review adheres to the guidelines outlined in the PRISMA statement~\cite{mallett2012benefits}. As shown in Figure~\ref{fig:11}, the PRISMA flowchart illustrates a systematic way of selecting papers. Initially, the search keywords were structured around three key aspects: data sources, industrial systems, and PHM/AI-related issues, with a focus on PHM related research using industrial open-source data from recent PHM data challenges. The search scope was restricted to articles published between 2018 and 2023 across Google Scholar, IEEE Xplore, Web of Science, Scopus, and ScienceDirect. Then, the literature search was conducted on December 18, 2023, using the predefined keywords. Subsequently, the identified records from five databases were consolidated, and duplicates were removed. After that, based on the exclusion criteria provided in Table~\ref{table:10}, reviewers further reviewed and evaluated the remaining articles and finally identified 59 representative papers for analysis, as presented in this survey.

\begin{table}[htbp]
\begin{center}
\caption{Exclusion Criteria for Screening Stage}
\label{table:10}
\renewcommand{\arraystretch}{1.5}
\begin{tabular}{|ll|}
\hline
\multicolumn{2}{|l|}{Exclusion criteria}                       \\ \hline
\multicolumn{1}{|l|}{E1} & Full text is not available          \\ \hline
\multicolumn{1}{|l|}{E2} & \begin{tabular}[c]{@{}l@{}}The method of the paper is not \\ based on machine learning techniques\end{tabular}      \\ \hline
\multicolumn{1}{|l|}{E3} & The paper is not written in English \\ \hline
\multicolumn{1}{|l|}{E4} & \begin{tabular}[c]{@{}l@{}}The research does not utilize \\ industrial open source data for analysis\end{tabular}      \\ \hline
\end{tabular}
\end{center}
\end{table}

\subsection{Overview of PHM Data Challenge Competitions}

\begin{table*}[htbp]
    \begin{center}
    \caption{Overview of PHM Data Challenge Competitions from 2018 to 2023}
    \label{table:1}
    \resizebox{\textwidth}{!}{
    \renewcommand{\arraystretch}{1.5}
    \begin{tabular}{c | c | c | c}
    \hline \hline
    \textbf{Competition}	& \textbf{Industrial Systems}  	& \textbf{Research Task}   & \textbf{No. of Papers}         \\ 
    \hline \hline
    2018 PHM NA	& Ion Mill Etching System	& Detection \& Diagnosis \& Assessment \& Prognosis & 12\\ 
    \hline
    2019 PHM NA & Fatigue Crack	& Assessment \& Prognosis & 4 \\ 
    \hline
    2020 PHM EU & Filtration System	& Prognosis & 8\\ 
    \hline
    2021 PHM EU & Manufacturing Production Line	&Detection \& Diagnosis & 7\\ 
    \hline
    2021 PHM NA	& Turbofan Engine & Prognosis & 8\\ 
    \hline
    2022 PHM EU	& Printed Circuit Board	& Detection \& Diagnosis & 6\\ 
    \hline
    2022 PHM NA	& Rock Drill & Detection \& Diagnosis & 5\\ 
    \hline
    2023 PHM AP	& Spacecraft Propulsion System & Detection \& Diagnosis & 5\\ 
    \hline
    2023 IEEE	& Gearbox & Detection \& Diagnosis & 4\\ 
    \hline \hline
    \end{tabular}}
    \end{center}
\end{table*}

From 2018 to 2023, PHM Society and IEEE Reliability Society have initiated nine PHM data challenge competitions which display challenges associated with analyzing industrial data across diverse industrial sectors. The topics of these competitions are extensive, encompassing different industrial systems such as Ion Mill Etching Tools, Filtration Systems, Manufacturing Production Lines, Turbofan Engines, Printed Circuit Boards, Rock Drills, Spacecraft Propulsion Systems, and Gearbox, among others. Furthermore, the range of problems posed covers the main tasks in PHM domain including detection, diagnosis, assessment, and prognosis. These tasks align with PHM's ultimate goal: to accurately evaluate and predict system health, degradation, and eventual failure, thereby improving system reliability, safety, and operational efficiency.

For ease of discussing different competitions, we've adopted a condensed naming convention for the PHM data challenge competition: "YEAR ORGANIZATION" indicates the respective year and organizer of the data challenge. Among the challenges discussed, one is organized by the IEEE Reliability Society, while the remaining eight are organized by the PHM Society. The PHM Society conducts an annual conference in North America, an Asia-Pacific conference in odd years, and a European conference in even years. We use the abbreviations "PHM NA", "PHM AP", and "PHM EU" to represent the competitions held in North America, Asia-Pacific, and Europe, respectively. For instance, "2018 PHM NA" refers to the challenge held in North America by the PHM Society in 2018 while "2023 IEEE" represents the competition hosted by the IEEE Reliability Society in 2023. Table~\ref{table:1} presents the systems, research tasks, and the number of research papers discussed in nine data competitions. For an in-depth overview of the PHM data challenge competitions and their associated datasets, please refer to Appendix and Section 3.


\subsection{Major Research Tasks within PHM}
By reviewing 9 PHM data challenge competitions, we summarized four major research tasks: detection, diagnosis, assessment, and prognosis.

\textbf{Detection}: It refers to identifying the presence of a fault, anomaly, or abnormal condition in a system or component. This is typically the first step in PHM, where sensors and monitoring systems are used to detect deviations from normal operations that might indicate a problem.

\textbf{Diagnosis}: Upon detecting anomalies, the diagnostic phase delves deeper to determine failure types or failure modes and find out the root causes of the problem. In diagnostic scenarios, detected failures often need to be classified into specific failure types.

\textbf{Assessment}: In the assessment phase, the current operational status and performance of the system is evaluated. Leveraging either historical data or recent machinery behavior, this phase evaluates potential risks or assesses the health status of the system in its present condition.

\textbf{Prognosis}: Prognosis leverages both current and historical data to forecast the future health of a system or the residual life of a machine. Commonly, this is referred to as predicting the Remaining Useful Life (RUL). This phase provides estimates on potential system or machinery failure timelines, thereby facilitating proactive maintenance planning. 

\begin{table*}[htbp]
    \begin{center}
    \caption{Overview of Detection and Diagnosis Problems in PHM Data Challenge Competitions}
    \label{table:3}
    \resizebox{\textwidth}{!}{
    \renewcommand{\arraystretch}{1.5}
    \begin{tabular}{| c | c | c | c | c | c |}
    \hline \hline
    \textbf{}	& \textbf{2021 PHM EU}  & \textbf{2022 PHM EU} & \textbf{2022 PHM NA} & \textbf{2023 PHM AP} & \textbf{2023 IEEE}               \\ 
    \hline \hline
    System	& Manufacturing Production Line	& Printed Circuit Boards & Rock Drill & Spacecraft Propulsion System & Gearbox \\ 
    \hline
    Failure Mode & 8 & 1+1+2 & 10 & 3 & 4 \\ 
    \hline
    Sensor Number & 50 & NA & 3 & 7 & 1 \\ 
    \hline
    Asset & NA & NA & 8 & 4 & NA \\ 
    \hline
    Operating Condition & 2 & 1+1+1 & 1 & 1 & 2 \\ 
    \hline
    Data Type & Time Series & Tabular & Pressure Signals & Pressure Signals & Vibration Signals \\ 
    \hline
    \multirow{2}{7em}{Volume of Data} & Limited & Medium & Large & Limited & Large \\
    & (70 Train, 29 Test) & (SPI:1924, AOI:1924) & (37229 Train, 16396 Test) & (177 Train, 46 Test) & (Total 50000) \\
    \hline
    Sampling Rate & 0.1 Hz & NA & 50 kHz & 1 kHz & 10 kHz \\ 
    \hline
    Sampling Interval & 1-3 hours & NA & NA & 1.2 s & 5 minutes \\ 
    \hline
    \end{tabular}}
    \end{center}
\end{table*}

\section{Methodology \& Analytics}
In Section 3, we categorize the competitions based on "Detection \& Diagnosis" task and "Assessment \& Prognosis" task and delve deeper into problems,  challenges, and ML method analysis in Section 3.1 and Section 3.2, respectively. In Section 3.3, we summarize a unified ML framework for the PHM domain.

\subsection{Detection \& Diagnosis}

In this subsection, we sequentially introduce the fundamental competition information, including background, objectives, challenges, and datasets, highlighting innovative approaches for fault detection and diagnosis problems. A summarization of the competition problems and their datasets is encapsulated in Table~\ref{table:3}. Furthermore, we provided a comprehensive summary of the methodologies utilized for detection and diagnosis problems, as outlined in Table~\ref{table:4}.

\subsubsection{2021 PHM EU (Manufacturing Production Line)}

In a collaborative effort with the Swiss Centre for Electronics and Microtechnology (CSEM), 2021 PHM EU offered a dataset derived from a real-world industrial manufacturing line dedicated to testing electrical fuses. The objective of this competition is to perform fault identification and classification, root cause analysis, and system operation parameter identification. The dataset showed eight unique system failure modes under two distinct operating conditions. The primary challenge with the training data was its class imbalance, as the majority of samples represented healthy conditions. Additionally, the dataset was quite small, with only 70 training samples and 29 testing samples. In the fuse test bench dataset, about 10\% of the data was missing, and it was not evenly distributed across variables.

Against this background, the winning solution was a combination of decision tree algorithms and a propagation system~\cite{de2021divide}. While the decision trees focused on diagnosis issue, the propagation system tackled chronology by incoperating a Kalman-style filter. To address data imbalance, the SMOTE (Synthetic Minority Oversampling Technique) method was utilized. Additionally,~\cite{ince2021fault} incorporated the Leave One Feature Out Importance (LOFO-Importance) package for capturing essential features. Subsequently, they applied linear discriminant analysis (LDA) for dimensionality reduction. During the modeling phase, different ML methods were employed — gradient boosting algorithms such as Extreme Gradient Boosting (XGBoost) and Light Gradient Boosting Machine (LightGBM), LDA classifier, and Gaussian process classifier. Especially for gradient boosting algorithms, they used Genetic Algorithms to optimize hyperparameters. A key observation was XGBoost's superior performance over the other algorithms. Differing from their approach,~\cite{aimiyekagbon2021rule} showcased a rule-based diagnostic technique, comparing its performance with that of decision trees and random forest methods. Additionally,~\cite{aydemir2021ensemble} proposed a regularized LSTM for sifting through vital features and then leveraged an ensemble of binary LSTM classifiers for fault detection and classification. 

Moreover, the XGBoost algorithm found favor not just in the aforementioned studies but also in three other distinct research papers~\cite{alfarizi2022extreme, ramezani2021explainable, tian2022high} outside the competition. The distinctions among these studies were as follows: For instance,~\cite{tian2022high} relied on a feature importance ranking (FIR) method, targeting enhanced performance and simplification in complex industrial classification scenarios, whereas~\cite{ramezani2021explainable} focused on the challenge of missing value imputation, harnessing Partial Least Squares (PLS-MV). A significant contribution of Ramezani’s framework was its explainability, achieved by pinpointing key performance indicators for each fault family using the SHAP method~\cite{lundberg2017unified}. 

\subsubsection{2022 PHM EU (Printed Circuit Boards)}
The 2022 European PHME Data Challenge, hosted in collaboration with Bitron Spa, focused on a classification issue within an actual industrial Printed Circuit Board (PCB) production line. The challenge's objectives contain three important tasks: (1) Task 1: predicting Automatic Optical Inspection (AOI) defect detection based on Solder Paste Inspection (SPI) data; (2) Task 2: predicting human-made visual inspection labels (OperatorLabel); (3) Task 3 predicting the human-assigned repair label (RepairLabel). Task 1 and Task 2 are binary classification problems, whereas Task 3 is a multi-class classification problem. This explains why, in Table~\ref{table:3}, the "Failure Mode" is listed as "1+1+2". In Table~\ref{table:3}, the "Operating Condition" is listed as "1+1+1", which means each task has its own operating condition. Meanwhile, the SPI dataset(2022 PHM EU) had missing information in specific fields and 95\% of the SPI data was classified as healthy, highlighting a significant imbalance issue.

In 2022 PHM EU, Gaffet’s team got 1st place using XGBoost method based on encoding and feature engineering~\cite{gaffet2022hierarchical}. Notably, they harnessed the SHAP method for model interpretation. Similarly,~\cite{taco2022novel} leveraged two tree-based algorithms (LightGBM and XGBoost). Their approach centered on solving classification problems by extracting significant statistical data during the feature engineering phase. The importance of feature engineering was further emphasized by~\cite{tang2022prediction}. They introduced a novel statistical feature extraction method coupled with a PinNumber-based technique. This method aimed to compress pin-level data into component-level information. When predicting automatic inspection defects, they integrated a neural network model, factoring in feeding imbalance control to navigate data imbalance challenges. Additionally, a random forest model was developed for both human inspection and repair predictions. Apart from feature engineering,~\cite{schmidt2022application} applied a multi-layer perceptron (MLP) neural network for defect predictions in automated inspections. For human inspection outcomes, a random forest algorithm was their choice, while decision trees were favored for predicting human repair labels. 

Outside of competition, Mirzaei's team delved deep into the challenge of imbalanced data. They proposed a data-level technique that integrated recursive feature elimination (RFE) for feature selection and oversampling methods. This ensured balanced representation for minority classes, enhancing the performance of multiple ML algorithms, from decision trees and random forests to SVMs and 1dCNNs~\cite{mirzaei2023application}. Additionally,~\cite{lee2022stream} introduced a new quality management paradigm termed Stream-of-Quality (SoQ) tailored for multi-stage manufacturing processes. By leveraging this dataset, they showcased the effectiveness of their methodology, offering promising avenues for refining industrial AI algorithms and methodologies in a systematic manner.

\subsubsection{2022 PHM NA (Rock Drill)}
Rock drills play a pivotal role in sectors like mining, tunneling, and construction. Due to the potential economic and human costs from work interruptions caused by rock drill faults, ensuring accurate fault diagnosis is necessary. 2022 PHM NA aimed to address fault classification problem under various product configurations. The main challenges is domain-shift problem, caused by data being collected from different rock drill machines. The dataset encompassed training data from five rock drill machines, validation data from another machine, and test data from two additional machines, highlighting potential domain shift issues between training, validation, and test data. This leads to heterogeneous signal distributions, which can negatively impact classification accuracy. The dataset~\cite{jakobsson2022dataset} encompasses ten distinct fault modes and one health mode, with training, validation, and testing data sizes of 34,045, 3,184, and 16,396, respectively. 

In the competition, ~\cite{oh2023hybrid} won the first place. They deployed a hybrid strategy, combining data-driven techniques with various signal-processing methods. Their approach harnessed domain adaptation, metric learning, and pseudo-label-based deep learning to construct an ensemble DL model for comprehensive fault classification. For samples that posed challenges for DL, they deployed signal processing methods like Dynamic Time Warping (DTW), and Cross-correlation, and used SVM for supervised learning. The runner-up solution introduced a data-cropping technique, employing a convolutional neural network (CNN) as a feature extractor to bridge data length discrepancies.~\cite{kim2023domain} innovatively addressed the domain-shift issue through a domain-adaptation-based scheme that harnessed a domain adversarial learning neural network for extracting domain invariant features while using maximum mean discrepancy (MMD) minimization for bridging distribution discrepancy, and a soft voting ensemble to reduce model uncertainty. Moreover, Minami's team proposed an ensemble approach, appending specialized models onto a baseline model. This ensemble incorporated domain adaptation strategies to accommodate domain fluctuations. The baseline model employed conventional ML algorithms like SVM, Random Forest, and XGBoost for whole multi-class classification, whereas the specialized sub-models, leveraging CNN for feature extraction and classification, concentrated on binary classifications targeting specific classes that poorly performed on the baseline model~\cite{minami2023novel}. 

\begin{table*}[htb]
\begin{center}
\caption{Overview of Detection and Diagnosis Methodologies in PHM Data Challenge Competitions}
\label{table:4}
\renewcommand{\arraystretch}{1.2}
\begin{tabular}{|l|lll|}
\hline
\multirow{2}{*}{} &
  \multicolumn{3}{c|}{\textbf{Methodology}} \\ \cline{2-4} 
 &
  \multicolumn{1}{l|}{\textbf{Deep Learning}} &
  \multicolumn{1}{l|}{\textbf{Conventional Machine Learning}} &
  \textbf{Feature Engineering} \\ \hline
\textbf{2021 PHM EU} &
  \multicolumn{1}{l|}{\begin{tabular}[c]{@{}l@{}}Regularized LSTM\\ Ensemble LSTM\end{tabular}} &
  \multicolumn{1}{l|}{\begin{tabular}[c]{@{}l@{}}XGBoost, Decision Trees\\ Random Forest\\ LDA, Rule-based Method\\ Gaussian Process\end{tabular}} &
  \begin{tabular}[c]{@{}l@{}}FIR\\ PLS-MV\\ FCM\\ SMOTE\end{tabular} \\ \hline
\textbf{2022 PHM EU} &
  \multicolumn{1}{l|}{1D-CNN} &
  \multicolumn{1}{l|}{\begin{tabular}[c]{@{}l@{}}Decision Trees, Random Forest\\ XGBoost, LightGBM\\ SVM, MLP\end{tabular}} &
  \begin{tabular}[c]{@{}l@{}}RFE\\ Statistical Feature Extraction\end{tabular} \\ \hline
\textbf{2022 PHM NA} &
  \multicolumn{1}{l|}{\begin{tabular}[c]{@{}l@{}}Domain Adaption\\ DANN, X-Vectors\\ Metric Learning\\ Pseudo Label Technique\end{tabular}} &
  \multicolumn{1}{l|}{\begin{tabular}[c]{@{}l@{}}Ensemble Learning\\ XGBoost\\ SVM\\ Deep Forest Algorithm\end{tabular}} &
  \begin{tabular}[c]{@{}l@{}}MMD\\ RFE\\ DTW\end{tabular} \\ \hline
\textbf{2023 PHM AP} &
  \multicolumn{1}{l|}{NA} &
  \multicolumn{1}{l|}{\begin{tabular}[c]{@{}l@{}}Similarity-based Method\\ K-means clustering, KNN\\ Decision Trees, XGBoost\\ Rule-based method\\ Ensemble Learning\end{tabular}} &
  \begin{tabular}[c]{@{}l@{}}Physical Feature Extraction\\ DTW\end{tabular} \\ \hline
\textbf{2023 IEEE} &
  \multicolumn{1}{l|}{\begin{tabular}[c]{@{}l@{}}ROCKET, LSTM-FCN\\ 1D-CNN with ResNet\\ Deep Residual Network\\ Residual-based CNN\end{tabular}} &
  \multicolumn{1}{l|}{Ensemble Learning} &
  \begin{tabular}[c]{@{}l@{}}Data Augmentation \\ Data Regularization\\ STFT\end{tabular} \\ \hline
\end{tabular}
\end{center}
\end{table*}

Outside the competition,~\cite{ling2023hydraulic} deployed an end-to-end fault classification framework derived from X-Vectors~\cite{snyder2018x}, achieving integrated optimization of both feature extraction and classification phases.~\cite{taco2023novel} proposed a novel deep forest algorithm that fused multi-grained scanning for feature extraction and a cascade forest for layered predictions to perform failure model classification.

\subsubsection{2023 PHM AP (Spacecraft Propulsion System)}
The Japan Aerospace Exploration Agency (JAXA) initiated a competition centered on the advancement of PHM technology for spacecraft propulsion systems. The primary objective was to accurately diagnose various states ranging from normal conditions to bubble anomalies, solenoid valve faults, and unknown faults. Simulation data was sourced from four distinct spacecraft, resulting in a dataset comprising 177 training samples and 46 test samples~\cite{2023PHMAPdata}. The paucity of data posed a challenge for data-driven approaches. 

In the competition, the champions~\cite{minami2023phm}, introduced a novel two-step approach. Initially, a similarity-based model was proposed for the categorization of data into four distinct states. Subsequently, for data corresponding to the solenoid valve fault, a model incorporating physic-inspired features was employed to pinpoint the fault location and estimate the valve opening ratio. They also deployed DTW~\cite{berndt1994using} on the training dataset which was instrumental in quantifying the variability across various segments of the sensor data. Standing in the second position,~\cite{lee2023hybrid} devised a hybrid approach combining the XGBoost-based method and the rule-based method. While the XGBoost-based approach primarily addressed comprehensive fault classification, the rule-based method was employed to formulate the solenoid valve opening ratio equation. This was accomplished through polynomial fitting, rooted in intrinsic physical characteristics, enabling a precise estimation of the solenoid valve opening ratio. 

Additionally,~\cite{kato2023anomaly} utilized the K-NN algorithms to classify faults and pinpoint the location of anomalies. Prior to the classification, a differentiation between normal and anomalous data was executed based on a similarity-based approach. Their estimation metrics for valve opening ratio were hinged upon the similarity of time series waveforms. Moreover,~\cite{aimiyekagbon2023expert} employed an ensemble framework, integrating K-means clustering and decision trees. This model, enriched by domain-specific expertise, exhibited good precision in both anomaly detection and fault diagnosis. Various approaches in 2023 PHM AP underscore the importance of similarity-based methods and the extraction of physical features when the dataset size is small. 

\begin{table*}[htbp]
    \begin{center}
    \caption{Overview of Assessment and Prognosis Problems in PHM Data Challenge Competitions}
    \label{table:6}
    \resizebox{\textwidth}{!}{
    \renewcommand{\arraystretch}{1.5}
    \begin{tabular}{| c | c | c | c | c |}
    \hline \hline
    \textbf{}	& \textbf{2018 PHM NA}  & \textbf{2019 PHM NA} & \textbf{2020 PHM EU} & \textbf{2021 PHM NA}               \\ 
    \hline \hline
    System	& Ion Mill Etching System & Fatigue Crack & Filtration System & Turbofan Engine \\ 
    \hline
    Failure Mode & 3 & 1 & 1 & 7 \\ 
    \hline
    Sensor Number & 5 & 2 & 3 & 14 \\ 
    \hline
    Asset & 20 & 8 & NA & NA \\ 
    \hline
    Operating Condition & Multiple & Variable Loading Conditions & 12 & 4 \\ 
    \hline
    Data Type & Time Series & Time Series & Time Series & Time Series \\ 
    \hline
    \multirow{2}{7em}{Volume of Data} & Limited Samples & Limited Samples & Limited Samples & Limited Samples \\
    & (20 Train, 5 Test) & (74 Train, 36 Test) & (24 Train, 8 Validation, 16 Test) & (90 Train Units, 38 Test Units) \\
    \hline
    Sampling Rate & 0.25 Hz & 5 Hz & 10 Hz & 1 Hz \\ 
    \hline
    Sampling Interval & Above 70 million seconds & 14000-100774 cycles & 200-350 s & 1-3 hours, 3-5 hours, Above 5 hours \\ 
    \hline
    \end{tabular}}
    \end{center}
\end{table*}

\subsubsection{2023 IEEE (Gearbox)}

The objective of 2023 IEEE is to develop ML-based models that can efficiently detect faults in the planetary gearboxes of industrial machinery using vibration signals. The dataset covers four prevalent sun gear faults: surface wear, chipped teeth, cracks, and missing teeth. Vibration signals have been recorded for a duration of five minutes each, with a sampling rate of 10 kHz, under two distinct operational conditions. Given the substantial dataset, comprising 50,000 samples, the deployment of DL techniques is feasible.

In 2023 IEEE.~\cite{lee2023ensemble} utilized ensemble-based CNN techniques to address fault detection via time-series vibration data. Their findings underscored robust performance of integrating three convolution kernel-based methods such as ROCKET (RandOm Convolutional KErnel Transform) method~\cite{dempster2020rocket}, one dimensional convolutional neural networks (1dCNN) integrated with ResNet, and a fusion of LSTM and Fully Convolutional Network (FCN) in delivering classification results for multivariate time-series data. Similarly, both~\cite{shen2023gear} and~\cite{kreuzer20231}, designed residual-based CNN models to address fault classification challenges. On the one hand,~\cite{shen2023gear} incorporated the Short-Time Fourier Transform (STFT)~\cite{huang2021tool}, transforming frequency domain signals into time-frequency domain signals using Fourier analysis - an useful tool for interpreting time-evolving signals. On the other hand,~\cite{kreuzer20231} leveraged data augmentation and regularization techniques, enabling model construction with fewer parameters without sacrificing performance quality. Additionally,~\cite{sobha2023comprehensive} harnessed a tree classifier to select valuable features from raw vibration signals, subsequently developing a sequential neural network model tailored for the concurrent detection of multiple gear faults.

\subsection{Assessment \& Prognosis}

Similar to the structure of subsection 3.1, we detail specifics of each competition, including the problems, challenges, and data-driven approaches applied. We also provide a consolidated summary of both the problems with their associated datasets and the proposed solutions in Table~\ref{table:6} and Table~\ref{table:7}, respectively.

\subsubsection{2018 PHM NA (Ion Mill Etching System)}
The 2018 PHM NA emphasized the analysis of fault behavior within the ion mill etch tool, tasking participants to develop a model from the sensor-derived time series data capable of accurately detecting, diagnosing, and prognosticating the time-to-failure for three principal failure modes—Flowcool leak (F1), Flowcool Pressure Too High Check Flowcool Pump (F2), and Flowcool Pressure Dropped Below Limit (F3). Additionally, the Ion Mill Etching (IME) dataset faced an imbalance issue, where the faulty data is much less than normal operation data.

For this dataset,~\cite{wu2021remaining,huang2018remaining,zhao2022deep,he2019failure}have each proposed strategies based on random forest algorithms for early degradation mode detection and diagnosis. Extending the exploration of ML techniques,~\cite{singh2018concurrent} evaluated an assortment of models, including Generalized Linear Models, MLP, Multivariate Adaptive Regression Splines, Support Vector Regression (SVR), random forest, etc. Among these, the random forest model distinguished itself with superior performance. Meanwhile,~\cite{zheng2021cross} utilized a knowledge distillation approach towards fault detection across various modes, aiming to improve detection performance of infrequent but critical faults.

Recent advancements in RUL prediction~\cite{kim2021challenges} have predominantly hinged on the application of DL algorithms to analyze complex multivariate time series data.~\cite{wu2021remaining,huang2018remaining,he2019failure} have all utilized Long Short-Term Memory (LSTM) networks and their variations, such as Gated Recurrent Units (GRU) and LSTM-based Metric Regression (LSTM-MR), to capture important features from the raw data.~\cite{hsu2022temporal} expanded upon this approach by integrating a Temporal Convolutional Network (TCN) with LSTM with attention mechanisms, which facilitated refined feature extraction from sensor data for accurate RUL prediction. Moreover, Liu et al.'s two-stage deep transfer learning framework aimed at achieving accurate RUL prediction. In the first stage, the developed model leveraged TCN for initial temporal feature learning, followed by domain adversarial learning for data alignment based on one fault mode. Then in the second stage, the first-stage model was fine-tuned based on other fault mode data to handle multiple fault modes and enhance the RUL prediction performance~\cite{liu2021two}. Distinct from these methods,~\cite{zhao2022deep} explored transformer networks, focusing solely on data from abnormal operation phases, while~\cite{lorenti2023predictive} provided a comprehensive comparison of state-of-the-art methods, including TCN, LSTM, attention-based mechanisms, CNN, and Transformer. Moreover,~\cite{singh2018concurrent} introduced a novel approach using DTW for RUL estimation, leveraging a library of truncated degradation curves and health score models to refine the final RUL predictions. 

\subsubsection{2019 PHM NA (Fatigue Crack)}
2019 PHM NA focused on the task of fatigue crack length estimation and prediction within aluminum structures at different points. The challenge harnessed wave signal data collected by piezoelectric sensors subjected to both static and dynamic tensile stresses. Given that the test dataset was limited to wave signals from several initial loading cycles, participants were challenged to estimate crack lengths where signal data were present and predict future crack growth for specified cycles lacking signal data. The complexity of the problem is that data-driven methodologies were viable when signals existed, whereas scenarios lacking wave data required exploration into physics-based strategies. Therefore, solutions derived from this challenge have predominantly employed a hybrid approach of data-driven and physics-based techniques.

Regarding the estimation of fatigue crack length with accessible wave signals, a variety of data-driven models and feature engineering strategies emerged, both within and beyond the competition's scope.~\cite{karimian2020neural} proposed a neural network architecture reliant on features manually derived from raw signals, such as the Pearson correlation coefficient, phase shifts, energy, and information entropy, to train models for accurate crack length estimation. Similarly,~\cite{kong2020hybrid} initiated their approach with signal preprocessing, utilizing techniques like band-pass filtering and phase alignment to mitigate noise and uncertainty before applying physically insightful feature extraction methods. Thereafter, they harnessed a random forest algorithm, optimizing it through feature selection and grid search for hyperparameter fine-tuning, to estimate crack lengths. Additionally,~\cite{youn2020fatigue} also leveraged a band-pass filter for feature extraction from raw wave signals, subsequently constructing an SVR model with hyperparameters optimized via grid search. Rao and collaborators, meanwhile, designed an ensemble learning regression model to improve estimation performance with four useful extracted features (root mean square value, correlation coefficient, first peak value, and the logarithm of kurtosis)~\cite{rao2020structure}.

Predictive modeling for scenarios lacking wave signal data required a pivot towards physics-based techniques, often in conjunction with data-driven insights, to forecast crack progression.~\cite{karimian2020neural} devised a Particle Filter (PF) strategy, integrating the Paris Law and outputs from the previously developed neural network as observational inputs to refine and update the crack propagation pathway.~\cite{kong2020hybrid} suggested an ensemble prognostics framework under consistent loading conditions, constructing a probability density function (PDF) for each instance. This was followed by a computation of weights derived from each PDF to output the final crack length prediction. Furthermore, when different loading conditions prevailed, Walker's equation was utilized to forecast crack lengths. Innovatively,~\cite{youn2020fatigue} introduced a trans-fitting approach, aimed at extracting the crack growth trend from training data and extrapolating it to the test data predictions. Additionally, Rao’s group advanced a variation version of Paris’ Law, aiming to elucidate the correlation between crack progression and the number of loading cycles~\cite{rao2020structure}.

\subsubsection{2020 PHM EU (Filtration System)}
\begin{table*}[htb]
\begin{center}
\caption{Overview of Assessment and Prognosis Methodologies in PHM Data  Challenge Competitions}
\label{table:7}
\renewcommand{\arraystretch}{1.2}
\begin{tabular}{|l|lll|}
\hline
\multirow{2}{*}{} &
  \multicolumn{3}{c|}{\textbf{Methodology}} \\ \cline{2-4} 
 &
  \multicolumn{1}{c|}{\textbf{Deep Learning}} &
  \multicolumn{1}{c|}{\textbf{Conventional Machine Learning}} &
  \multicolumn{1}{c|}{\textbf{Feature Engineering}} \\ \hline
\textbf{2018 PHM NA} &
  \multicolumn{1}{l|}{\begin{tabular}[c]{@{}l@{}}Transfer Learning\\ LSTM, LSTM-MR, GRU\\ Transformer\\ CNN\\ TCN, TCN-LSTM, TCN-DANN\end{tabular}} &
  \multicolumn{1}{l|}{\begin{tabular}[c]{@{}l@{}}Random Forest, Gradient Boosting\\ Logistic Regression\\ Generalized Linear Model\\ SVR, MLP\\ MASR\end{tabular}} &
  \begin{tabular}[c]{@{}l@{}}DTW\\ SVR-RFE\\ Degradation Curve Library\end{tabular} \\ \hline
\textbf{2019 PHM NA} &
  \multicolumn{1}{l|}{NA} &
  \multicolumn{1}{l|}{\begin{tabular}[c]{@{}l@{}}Random Forest, Gaussian Process\\ SVR \\ Neural Network\\ Ensemble Learning\\ Linear Regression\end{tabular}} &
  \begin{tabular}[c]{@{}l@{}}Grid Search\\ Genetic Algorithm\\ Paris's Law\\ Walker's Equation\\ Particle Filter\end{tabular} \\ \hline
\textbf{2020 PHM EU} &
  \multicolumn{1}{l|}{\begin{tabular}[c]{@{}l@{}}Neural Turing Machine\\ Transfer Ensemble Learning\\ LSTM, Bi-LSTM, TEL-Bi-LSTM\\ Autoencoder-Regression Network\\ Deep CNN\end{tabular}} &
  \multicolumn{1}{l|}{\begin{tabular}[c]{@{}l@{}}Random Forest, Gradient Boosting\\ Gaussian Process\\ Kernel Regression\\ SVR\\ Ensemble Learning\end{tabular}} &
  \begin{tabular}[c]{@{}l@{}}Simple Statistics Method\\ Linear SVM Coefficient\\ Correlation Metric\\ RFE, Health Index\\ Monotonicity Test\end{tabular} \\ \hline
\textbf{2021 PHM NA} &
  \multicolumn{1}{l|}{\begin{tabular}[c]{@{}l@{}}Deep CNN, FCN, VGG\\ ResNet(Residual Block)\\ GoogLeNet(Inception Module)\end{tabular}} &
  \multicolumn{1}{l|}{\begin{tabular}[c]{@{}l@{}}Random Forest, XGBoost\\ Extreme Random Forest\\ ANN\end{tabular}} &
  \begin{tabular}[c]{@{}l@{}}PCA\\ XAI,SHAP, LIME\end{tabular} \\ \hline
\end{tabular}
\end{center}
\end{table*}

In the domain of industrial maintenance, filtration systems are crucial for helping process pollutants from industrial equipment, ensuring seamless system operation. A predominant challenge encountered within these systems is filter clogging - a phenomenon where accumulated pollutants impede flow rates, thereby disrupting standard industrial workflows. To address this issue, 2020 PHM EU concentrated on the prediction of filtration systems' RUL. RUL, in this context, is delineated as the time until the pressure differential across the filter breaches a threshold of 20 psi. The challenge provided the PHME20 dataset, collected from a controlled experimental setup designed to simulate filter clogging at various contamination levels. Twelve distinct conditions were established influenced by two operational parameters: solid ratio (\%) and particle size (µm). Additionally, the 2020 PHM EU dataset demonstrated domain shift problem, with training data featuring small and large particle sizes, while the test data included medium particle sizes.
          
The champion solution~\cite{lomowski2020method} employed a novel hybrid approach, combining kernel regression with fundamental statistical methodologies. Meanwhile, the runner-up,~\cite{beirami2020data}, adopted different cutting-edge feature engineering, and ML methods. Their process began with the extraction of features through a rolling window technique and proceeded with feature selection informed by the linear kernel Support Vector Machine (SVM) coefficients, RFE, correlation matrix, and monotonicity testing. A four-layer sequential neural network was their model of choice, supported by K-fold cross-validation throughout the training and validation stages. Capturing the third position,~\cite{ince2020remaining} conducted a comparative study of tree-based algorithms and the Bayesian approach. Specifically, random forest, gradient boosting, and Gaussian process regression were utilized to estimate the RUL of the filtration system, supplemented by a novel fault-based RUL assignment that integrated "Piecewise RUL Assignment" and "Linear RUL Assignment".

\begin{figure*}[!h]
\centering\includegraphics[width=\linewidth]{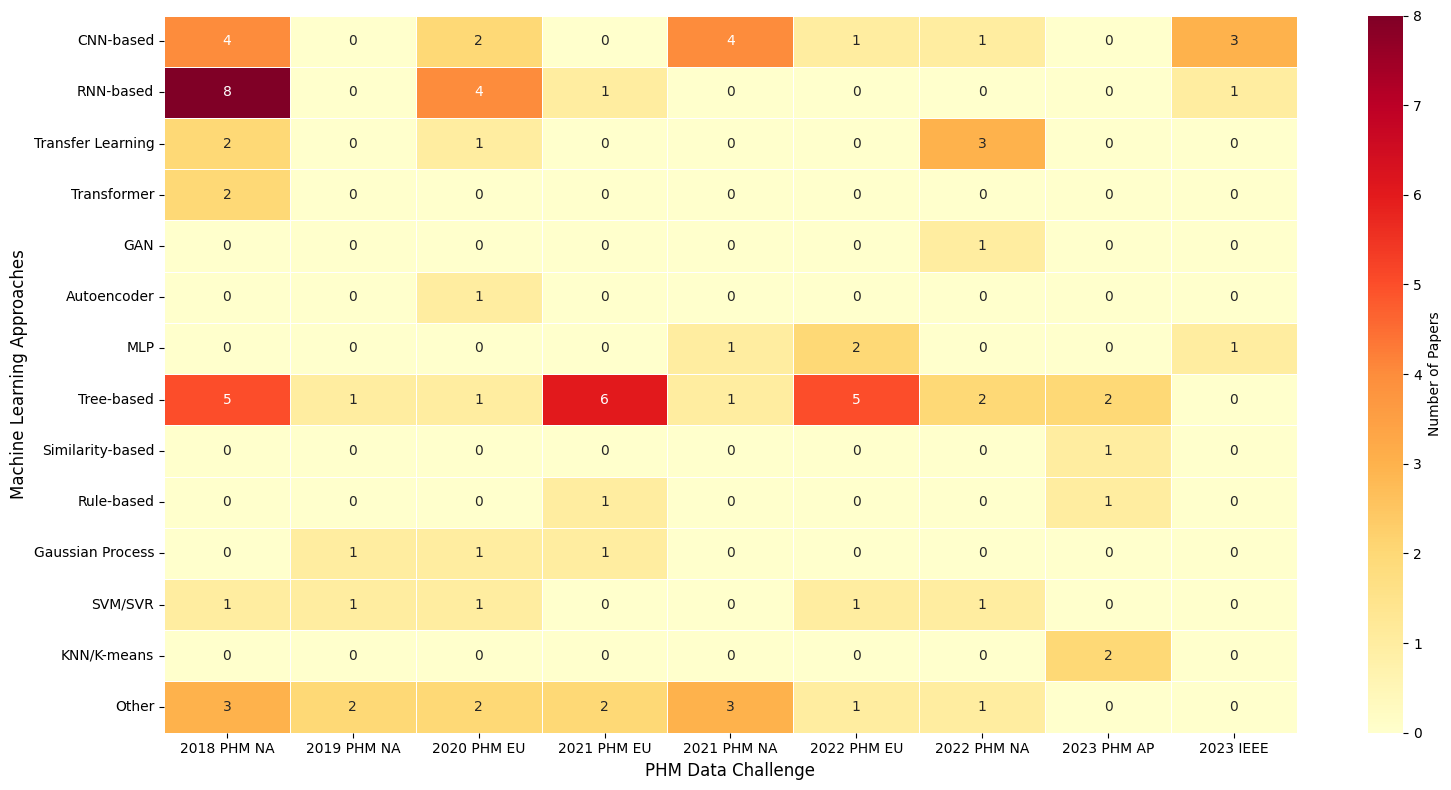}
\caption{Distribution of ML Approaches Across PHM Data Challenge Competitions}
\label{fig:2}
\end{figure*}

Contrasting with the feature engineering and conventional ML strategies, some researchers explored some DL methods.~\cite{vu2021deep} introduced a CNN-based DL methodology for RUL prediction, marked by two architectural innovations: a Parameterized Fully Connected Layer that adjusts network weights in response to operational parameter shifts, and a multi-head predictor tailored to distinct degradation process stages. Additionally,~\cite{tian2023novel} presented a new transfer ensemble learning (TEL) framework, leveraging metric learning alongside domain dissimilarity metric and Kullback–Leibler (KL) divergence, to enhance model generalization from source to target domains. This TEL framework amalgamated with a bidirectional long short-term memory (Bi-LSTM) algorithm, coined as TEL-Bi-LSTM, was offered for RUL estimation under different operating conditions. In another innovative approach,~\cite{ince2023joint} proposed a joint autoencoder-regression network, a deep neural architecture that fused a CNN autoencoder with an LSTM network regressor in an end-to-end training paradigm. Genetic algorithms were instrumental in optimizing hyperparameters for this architecture. Additionally, Lee’s team developed a distinct strategy by first establishing a health assessment criterion~\cite{lee2021data}. They defined a Health Index (HI) for the filter system and utilized K-means clustering for the categorization of the system's health stages. Subsequent HI predictions were facilitated by the Bi-LSTM algorithm, thus determining the system's RUL. Lastly, Falcon et al. introduced an innovative sequence modeling technique termed the Neural Turing Machine (NTM)~\cite{falcon2022neural}. Conceptualized as a computational architecture, the NTM leverages available data to interact with an external memory component, an approach that facilitates enhanced accuracy in predictions. This model stands out for its ability to generate more precise outcomes when benchmarked against the prevalent LSTM-based solutions that dominate the field.

\subsubsection{2021 PHM NA (Turbofan Engine)}
2021 PHM NA was primarily focused on the prediction of RUL for turbofan engines~\cite{chao2021phm}, specifically under four distinct flight conditions and seven failure modes. Participants were required to create predictive models leveraging the N-CMAPSS dataset, aiming to accurately forecast RUL using complex condition monitoring data. The dataset, a collection consisting of 90 synthetic run-to-failure trajectories for training and an additional 38 truncated datasets for testing, served as a comprehensive foundation for developing robust predictive algorithms to predict RUL accurately.

During the competition, the winning team, led by Lovberg, put forward an innovative approach leveraging a deep convolutional neural network (DCNN)~\cite{lovberg2021remaining}. This network was distinguished by its use of dilated convolutions complemented by gated linear unit activations and integration of residual skip connections. Those techniques were designed to expand the network's receptive field and enhance flexibility, so as to reduce the complexity of the neural network architecture by using less number of parameters, but still having comparable performance. Additionally, they adopted a strategic sequence sampling method, minimizing less informative samples while retaining enough degradation signals for the network's input.  Moreover, Solis-Martin et al. and DeVol’s team pursued advancements in DCNNs as well. The former developed a two-level DCNN system where the first-level DCNN focused on extracting important features from raw data and the second-level DCNN leverages the output from the first level to accurately estimate the RUL~\cite{solis2021stacked}. The latter drew upon established DL architectures, deploying the basic blocks or modules from the VGG, GoogLeNet, and ResNet designs into their DCNN framework. This enabled a comparative analysis of model performances using a variety of well-known architectural features~\cite{devol2021inception,devol2022evaluating}.

Outside of the competition,~\cite{cohen2023fault} addressed concerns regarding the uncertain and poor interpretability of deep learning models by integrating principal component analysis (PCA) to refine time-domain feature sets and subsequently applying four supervised learning techniques, including artificial neural network, random forest, extreme random forest and XGBoost, to estimate RUL. Their innovative use of a custom loss function in conjunction with traditional ANN models got the best results in both Area Under the Receiver Operating Characteristic (AUROC) and Area Under the Precision-Recall (AUPR) metrics. Moreover, the domain of PHM has witnessed an upsurge in the application of XAI techniques to enhance the interpretability and trustworthiness of ML models. Various methods, including LIME, SHAP, LRP, Image-Specific Class Saliency Maps, and Gradient weighted Class Activation Mapping (Grad-CAM), have been reported in recent literature~\cite{cohen2023trust,cohen2023shapley,solis2023soundness}. These techniques are useful in elucidating the decision-making processes of complex models.

\subsection{Comprehensive Summarization of Data-Driven Approaches in Recent PHM Data Challenge Competitions}

\begin{figure}[h!]
\centering\includegraphics[width=\linewidth]{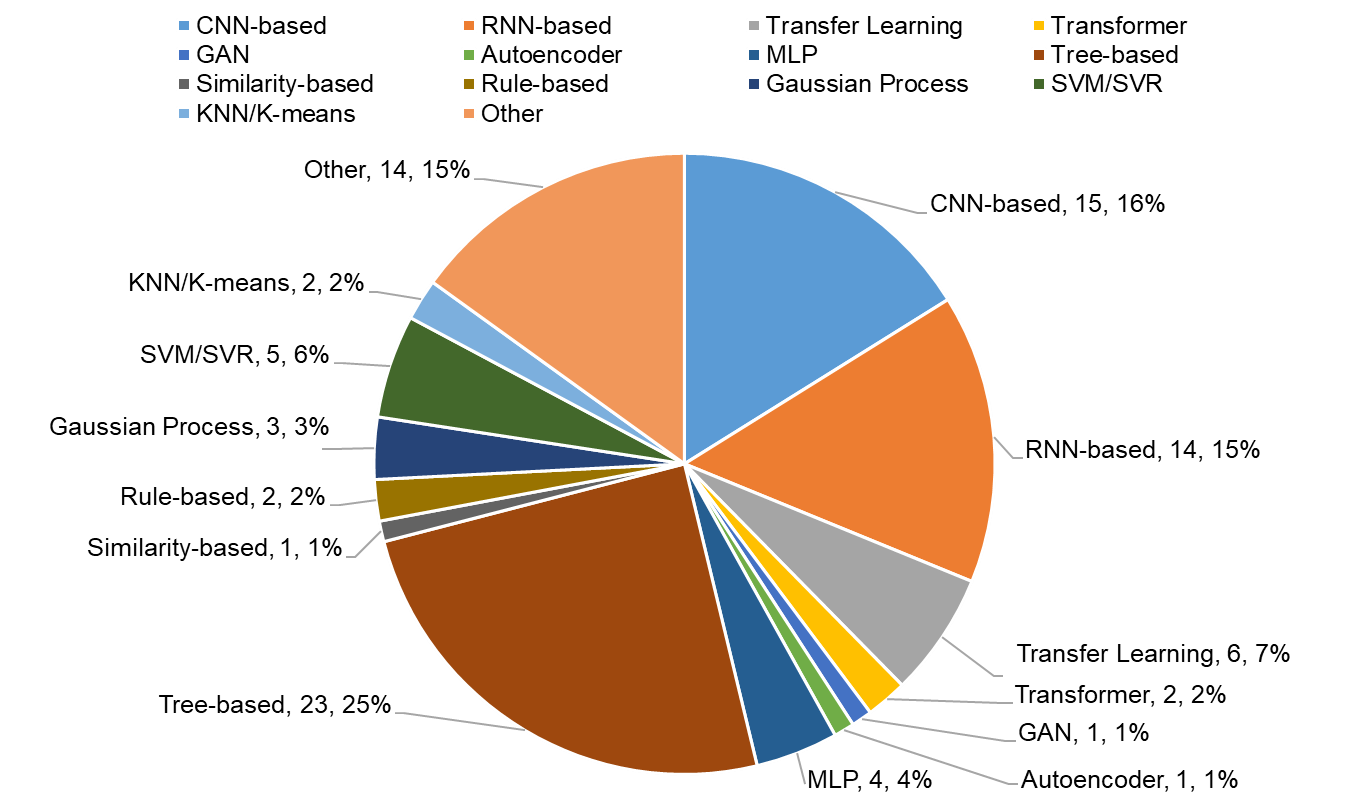}
\caption{The Density of Various ML Approaches in PHM Data Challenge Competitions from 2018 until December 2023}
\label{fig:3}
\end{figure}

After conducting an in-depth review of data competitions from the last six years, Figure~\ref{fig:2} provides details into the publication count and the frequency of particular ML approaches within each competition. Furthermore, Figure~\ref{fig:3} illustrates the density of specific ML or DL approaches discussed in this paper. It is important to note that we count the occurrences of distinct ML methods mentioned in a research paper. Given that some articles employ multiple ML methods, the aggregate count of methods exceeds the total number of publications.

\begin{figure*}[h!]
\centering\includegraphics[width=\linewidth]{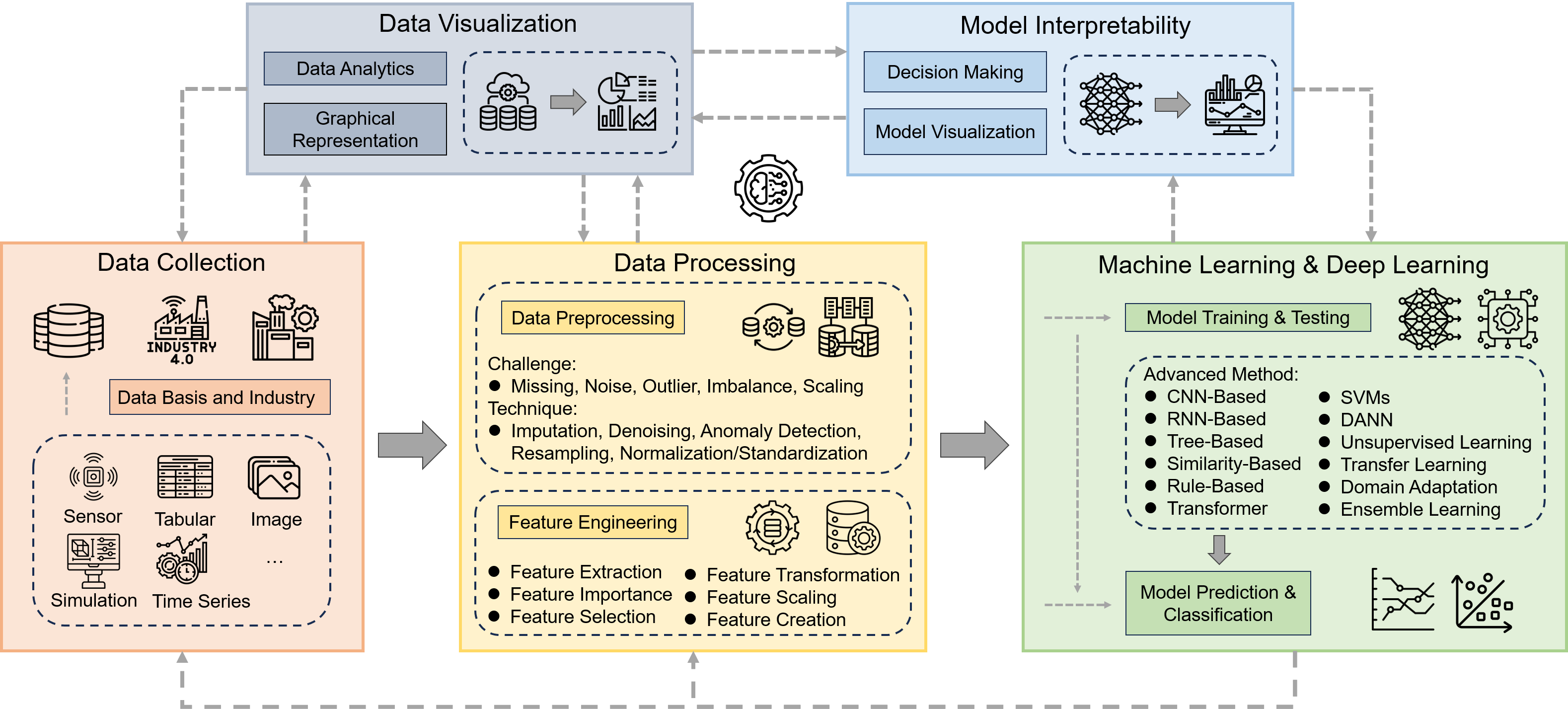}
\caption{A Unified Machine Learning Framework in Prognostics and Health Management Domain}
\label{fig:4}
\end{figure*}

Moreover, we have summarized a unified ML framework that concludes ML approaches for PHM data competitions during this period. As shown in Figure~\ref{fig:4}, it encompasses five primary components: Data Collection, Data Processing, Data Visualization, Conventional Machine Learning \& Deep Learning, and Model Interpretability.

\textbf{Data Collection}:
Data collection is the foundation of these open-source data challenge competitions (PHM Society and IEEE Reliability Society). Companies across various industries contribute datasets to encourage PHM community to develop innovative solutions to address real-world challenges. Typically, these industrial datasets encompass a diverse range of data types, including but not limited to time-series data, tabular data, images, sensor readings, and simulation data.

\textbf{Data Processing}:
An important stage before ML model development is data processing, which is crucial for enhancing data quality and ensuring effective model training~\cite{cofre2021big,tang2020data,correa2022data,griffiths2022managing}. This stage can be further subdivided into two aspects: data preprocessing and feature engineering. Data preprocessing addresses raw data challenges, typically including missing data, noise, outliers, data imbalance, and scaling issues. Techniques like imputation~\cite{eekhout2012missing}, denoising~\cite{de2021divide}, outlier detection~\cite{marti2018effects}, resampling~\cite{cicak2023handling}, and normalization \& standardization~\cite{lecun2015deep,goodfellow2016deep} are commonly applied. Feature engineering follows, refining data representation post-preprocessing. This step often uncovers hidden insights and deepens understanding of data, therefore significantly enhancing model performance and predictive capabilities in the later stage~\cite{sim2020tutorial}.

\textbf{Data Visualization}:
Data visualization is another important aspect of the ML process. It involves transforming data into intuitive graphical representations, such as graphs or charts. Researchers can use these charts to better observe trends, patterns, or outliers in data, which further help people have a better understanding of data and generate some useful insights. In the PHM domain, effective visualization can help initial data exploration and analysis, and accelerate data processing and ML modeling development~\cite{carley2022data,cheng2022systematic}. 

\textbf{Conventional Machine Learning \& Deep Learning Techniques}:
Post data processing, various conventional ML, and DL methods are developed to solve the competition problems using useful and usable data, as shown in Figure~\ref{fig:4}. This segment includes "model training and testing" and "model prediction and classification". During the training and testing phase. ML/DL techniques are diverse, including Tree-Based Methods, CNNs, RNNs and their variants, Transformers, DANN,  SVMs, Unsupervised Learning, Ensemble Learning, Similarity-Based Methods, Rule-Based Methods, Transfer Learning, Domain Adaptation, etc.~\cite{serradilla2022deep,qiu2023deep} Once the ML models are trained and tested well, they could be deployed into various PHM applications which include detecting abnormal performance or faults, classifying different faults, assessing the current health state of systems, or predicting the RUL of components. 

\textbf{Model Interpretability}:
Except for the pursuit of accuracy in prediction and classification problems, model interpretability is becoming more and more important. Model interpretability refers to interpreting ML model outputs~\cite{molnar2020interpretable} which helps humans understand the 'why' behind a model's predictions, facilitating collaboration and more informed decision making. Many explainable artificial intelligence (XAI) methods such as Local Interpretable Model-Agnostic Explanations (LIME)~\cite{ribeiro2016should} and SHapley Additive exPlanations (SHAP)~\cite{lundberg2017unified} are utilized in the PHM data challenge competitions to explain model outputs in PHM~\cite{solis2023soundness}. 

\section{Challenges and Possible Solutions}
In this section, we summarize common challenges regarding data-related issues and model-related issues and analyze relevant solutions applied to solve these challenges, shown in Table~\ref{table:9}. Data-related challenges encompass issues such as missing data, data imbalance, and domain shift while model-related challenges include the critical aspects of model selection, interpretability of ML models, and their robustness and generalization capabilities. These challenges highlight the complexities of developing effective ML methods to solve various PHM problems. 

Moreover, the limitations of current PHM competitions reveal a gap in adopting systematic approaches for building effective PHM systems, and a need for multi-modal machine learning analysis. We also suggest further steps, aiming to accelerate the development of next-generation ML-driven PHM solutions.

\begin{table*}[htb]
\begin{center}
\caption{Summary of Common Challenges Regrading Data-related and Model-related Issues}
\label{table:9}
\renewcommand{\arraystretch}{1.2}
\begin{tabular}{|ll|l|l|}
\hline
\multicolumn{2}{|l|}{Common Challenges} &
  Potential Solutions &
  Related Competitions \\ \hline
\multicolumn{1}{|l|}{} &
  Missing Data &
  \begin{tabular}[c]{@{}l@{}}(1) Listwise or Pairwise Deletion\\ (2) Imputation (LOCF, PLS-MV)\\ (3) Other\end{tabular} &
  \begin{tabular}[c]{@{}l@{}}2021 PHM EU\\ 2022 PHM EU\end{tabular} \\ \cline{2-4} 
\multicolumn{1}{|l|}{Data Issue} &
  Data Imbalance &
  \begin{tabular}[c]{@{}l@{}}(1) Resampling\\ (Oversampling or Downsampling)\\ (2) Synthetic Data Generation\\ (SMOTE and Variants)\\ (3) Transfer Learning Techniques\end{tabular} &
  \begin{tabular}[c]{@{}l@{}}2018 PHM NA\\ 2021 PHM EU\\ 2022 PHM EU\end{tabular} \\ \cline{2-4} 
\multicolumn{1}{|l|}{} &
  Domain Shift &
  \begin{tabular}[c]{@{}l@{}}(1) Transfer Learning Techniques\\ (2) Domain Adaptation Techniques\\ (3) DANN\end{tabular} &
  \begin{tabular}[c]{@{}l@{}}2020 PHM EU\\ 2022 PHM NA\end{tabular} \\ \hline
\multicolumn{1}{|l|}{} &
  Model Selection &
  \begin{tabular}[c]{@{}l@{}}(1) Need to Consider Volume and Quality \\ of Data\\ (2) A Trade-off Between Computational\\ Cost and Performance\end{tabular} &
  All Competitions \\ \cline{2-4} 
\multicolumn{1}{|l|}{Model Issue} &
  Model Interpretability &
  \begin{tabular}[c]{@{}l@{}}(1) Explainable Artificial Intelligence\\ (XAI) methods like LIME, SHAP, etc.\end{tabular} &
  \begin{tabular}[c]{@{}l@{}}No Competition \\ Required\end{tabular} \\ \cline{2-4} 
\multicolumn{1}{|l|}{} &
  \begin{tabular}[c]{@{}l@{}}Model Robustness \\ \& Generalization\end{tabular} &
  \begin{tabular}[c]{@{}l@{}}(1) Data Augmentation\\ (2) Regularization Techniques\\ (3) Ensemble Methods\\ (4) Transfer Learning \& Domain Adaptation\end{tabular} &
  All Competitions \\ \hline
\end{tabular}
\end{center}
\end{table*}

\subsection{Data-related Issues}
\subsubsection{Missing Data}
Handling missing data in ML, especially in the context of the PHM is necessary, as incomplete data can significantly impact the model performance and prediction accuracy. The strategies to address missing data in PHM have evolved, encompassing a range of techniques from basic deletion to advanced imputation methods. A straightforward strategy is the deletion of data points with missing values, such as listwise or pairwise deletion. For example, in the 2022 SPI dataset, instances having missing values in crucial identifiers like "Panel\_ID", "Figure\_ID", and "Component\_ID" were eliminated~\cite{taco2022novel}. However, this approach can lead to the loss of valuable information to some degree. Regarding imputation methods, several innovative methods were used in 2021 PHM EU.~\cite{de2021divide} applied the Last Observation Carried Forward (LOCF) method, coupled with backward filling, to address the gaps, while  ~\cite{ramezani2021explainable} introduced a novel approach, PLS-MV, a partial least squares-based method for imputing missing values. Interpolation was another technique used to estimate missing values, ensuring that cases with absent data did not skew the results. In addition to various imputation methods highlighted in the competitions, there remains scope for exploration in future research such as mean/median/multiple, K-Nearest Neighbors(KNN), regression imputation, maximum likelihood estimation, and ML-based approaches~\cite{huang2022reliable}. 

\subsubsection{Data Imbalance}
Data imbalance arises when the distribution of classes in a dataset is uneven. This issue is evident in PHM due to the rarity of failure events in comparison to data representing normal conditions. Data imbalance can significantly undermine the performance of ML models, particularly in classification tasks. When trained on imbalanced data, models may become biased towards the majority class (normal operation) and may not effectively recognize the minority class (failure). This leads to poor performance in predicting failures, which may bring ineffective maintenance planning and unexpected downtimes. 

To counteract data imbalance in PHM, various strategies have been adopted. Resampling techniques are commonly used to adjust the dataset to balance the class distribution, either by oversampling the minority class or downsampling the majority class. Synthetic data generation is another approach, as demonstrated by a team in 2021 PHM EU that used the SMOTE to create synthetic samples of the minority class to balance the dataset~\cite{de2021divide}. Advanced algorithms also play a crucial role in addressing data imbalance. For instance, Tang's team in 2022 PHM EU used the FIR to control the imbalance ratio in each mini-batch during neural network training~\cite{tang2022prediction}. Liu et al. employed transfer learning and domain adaptation techniques in 2018 PHM NA to facilitate knowledge transfer across different fault modes, addressing the issue of insufficient data in specific faults~\cite{liu2021two}. Moreover, during the 2018 PHM NA competition, many teams utilized Random Forest, an ensemble learning method known for its proficiency in handling imbalanced data, by constructing a forest of decision trees. 

Beyond the techniques showcased in the competitions, new DL approaches are proposed, such as the semi-supervised information maximizing generative adversarial network~\cite{wu2020ss}, the integration of deep residual networks with auxiliary classifier generative adversarial networks~\cite{chen2022fault}, the combination of DL with SMOTE~\cite{dablain2022deepsmote}, etc. Furthermore, a standardized experimental framework is proposed by Aguiar's team in order to evaluate 24 state-of-the-art data stream algorithms across 515 imbalanced data streams ~\cite{aguiar2023survey}. Going forward, there is a need for the exploration of additional DL-based strategies to enhance the handling of data imbalance in the PHM domain.

\subsubsection{Domain Shift}
Domain shift refers to the changes in the data distribution between the training phase (source domain) and the real-world application phase (target domain) of the ML models. This phenomenon frequently occurs when models, initially trained on data from a specific set of machines or under certain conditions, are subsequently applied to different machines or varied operating conditions. Such a shift can markedly affect the performance and reliability of ML models in PHM, making tackling with domain shift problem essential for maintaining the robustness and effectiveness of PHM systems. Transfer learning has emerged as a primary solution to domain shift challenges, as it can achieve the adaptation of models from one domain to be effective in another through fine-tuning and domain adaptation techniques. For instance, Kim et al. utilized domain adversarial neural networks (DANN) along with the minimization of MMD to tackle domain discrepancies in 2022 PHM NA~\cite{kim2023domain}. Meanwhile, Oh et al. employed the deep CORAL method, calculating coral loss to diminish domain discrepancy effects by aligning the covariance of the six training domains with that of the test domain~\cite{oh2023hybrid}. Moreover, Tian's team developed a novel TEL framework, facilitating knowledge transfer from the source to the target domain in 2020 PHM EU~\cite{tian2023novel}. Additionally, robust modeling approaches and the ability to continuously update models with new data are also useful and imperative for mitigating the impact of domain shift.

\subsection{Model-related Issues}
\subsubsection{Model Selection: Conventional Machine Learning or Deep Learning?}
In PHM data challenge competitions, participants often grapple with the difficult decision of whether to use conventional ML algorithms or advanced DL models. This choice is influenced by various factors: the volume and quality of available data, and a trade-off between computational cost and performance. An analysis of research papers from the past six years reveals insightful trends and preferences in model selection:

\textbf{(1) The volume and quality of training data play a significant role in determining the choice of modeling approach.} Research teams often choose DL for building data-driven models when training data are abundant. Conversely, in scenarios with limited training data, conventional ML methods, augmented by feature engineering, are more commonly employed. 

This trend is clear when addressing classification problems within PHM competitions. As depicted in Table~\ref{table:4} and Figure~\ref{fig:2}, 2022 PHM NA and 2023 IEEE competitions, because of their large datasets, have facilitated the application of DL techniques. In 2023 IEEE, various deep CNN models are proposed, including ensemble-based CNNs~\cite{lee2023ensemble} and residual-based CNNs~\cite{shen2023gear,kreuzer20231}. Additionally, in 2022 PHM NA, many teams integrate diverse DL approaches with transfer learning such as metric learning + pseudo label-based DL~\cite{oh2023hybrid}, CNN + DANN~\cite{kim2023domain}, X-Vectors~\cite{ling2023hydraulic}. However, the scenario differed for the 2022 PHM EU competition, which provided approximately 2000 samples and encountered data imbalance challenges. Given these constraints, many teams leaned towards tree-based methods known for their robustness and ability to address data imbalance, deploying algorithms like XGBoost~\cite{gaffet2022hierarchical,taco2022novel}, LightGBM~\cite{taco2022novel}, decision trees~\cite{mirzaei2023application}, and random forest~\cite{tang2022prediction,mirzaei2023application}. In competitions with significantly fewer data samples, such as 2021 PHM EU and 2023 PHM AP, where the dataset sizes were around 100-200 samples, the deployment of DL was impractical due to its requirement for a large volume of data. Instead, conventional ML methods proved more effective. In 2021 PHM EU, tree-based methods were predominantly used~\cite{de2021divide,ince2021fault,alfarizi2022extreme}, while in 2023 PHM AP, similarity-based~\cite{minami2023phm,kato2023anomaly}, rule-based~\cite{lee2023hybrid}, KNN/K-means~\cite{kato2023anomaly,aimiyekagbon2023expert}, and tree-based methods~\cite{lee2023hybrid,aimiyekagbon2023expert} are proposed, emphasizing the adaptability of traditional methods to limited data scenarios.

In the realm of RUL prediction tasks, the inherent long-time series nature of the datasets, even when datasets are smaller (less than 100 samples), allows for different approaches to data utilization and data augmentation. By considering each time point or a sequence of time points (data window) as an independent training sample, the effective size of the dataset can be substantially increased. This is evident from Table~\ref{table:7} and Figure~\ref{fig:2}, which indicate the application of both DL and conventional ML methods across various competitions such as  CNN-based~\cite{vu2021deep,solis2021stacked,devol2021inception}, RNN-based~\cite{wu2021remaining,tian2023novel}, transfer learning~\cite{liu2021two,tian2023novel}, Autoencoder-based~\cite{ince2023joint}, transformer-based~\cite{zhao2022deep,lorenti2023predictive} methods, and conventional ML with physics-based approaches~\cite{kong2020hybrid,youn2020fatigue} to tackle RUL prediction challenges. 

\textbf{(2) When selecting DL models, there is often a trade-off between computational cost and performance.} In the competitions, participants may choose highly complex DL models that require substantial computational resources to achieve even a small improvement in accuracy. While this approach might secure a higher position on the leaderboard, it may not be the most practical choice for real-world implementation. In practice, the value of such a small accuracy improvement needs to be weighed against the increased computational cost and potential scalability issues. 

\textbf{(3) When datasets become publicly accessible, there is a clear shift towards the adoption of more sophisticated DL approaches in PHM data challenge competitions over time.} Taking the 2018 PHM NA competition as a case study, our review of 12 published papers utilizing this dataset reveals that the methodologies initially favored by the competition teams largely comprised tree-based methods, SVMs, and basic LSTM algorithms~\cite{huang2018remaining,singh2018concurrent}. However, as time goes by, there is a discernible trend towards the development of more complex DL algorithms. This includes but is not limited to, GRU~\cite{wu2021remaining}, knowledge distillation~\cite{zheng2021cross}, a two-stage deep transfer learning framework utilizing TCN and DANN~\cite{liu2021two}, TCN combined with LSTM~\cite{hsu2022temporal}, attention mechanisms~\cite{hsu2022temporal,lorenti2023predictive}, and transformers~\cite{zhao2022deep,lorenti2023predictive}. This trend underscores a growing need for continuous innovation and refinement of DL methods in the PHM domain.

\subsubsection{Machine Learning Model Interpretability}

In PHM data challenge competitions, accuracy in prediction and classification is still the primary goal. However, with the advance of ML and DL, emphasizing the importance of model interpretability is becoming as crucial as their accuracy in prediction and classification, because it helps to illustrate the model decision-making process and plays an important role in error analysis and further model refinement in PHM.

While only a few teams in competitions have used model interpretability methods to explain the output of their models, outside of competitions, various techniques have been developed and applied to enhance ML model interpretability~\cite{linardatos2020explainable,arrieta2020explainable}. As the complexity of deep neural network models, often referred to as "black boxes", some previous research utilized interpretable methods like Layer-wise Relevance Propagation (LRP), Gradient-weighted Class Activation Mapping in CNNs, and attention mechanisms in sequential models to shed light on model decision-making processes~\cite{solis2023soundness}. Additionally, model agnostic methods like SHAP~\cite{lundberg2017unified,sundararajan2020many} and LIME~\cite{ribeiro2016should}have been instrumental in offering insights into model behavior. For instance, Baptista et al. applied the SHAP model to evaluate the outcomes of three different algorithms (Linear Regression, MLP, and Echo State Network) using the Commercial Modular Aero-Propulsion System Simulation (C-MAPSS) dataset (jet engines)~\cite{baptista2022relation}. Moreover, Moradi et al. introduced an interpretable artificial neural network designed for the automatic selection and fusion of features to develop optimal health indicators from data gathered through structural health monitoring (SHM)~\cite{moradi2022interpretable}. Furthermore, in an era increasingly focused on ethical and responsible AI, transparent and interpretable models are key to not only enhancing the technical aspects of AI solutions in PHM but also extending to ensuring their successful integration and acceptance in real-world applications~\cite{vollert2021interpretable}.

\subsubsection{Model Robustness and Generalization}

Model robustness is defined as the ability of a model to maintain its performance when facing diverse challenges like noise, outliers, and adversarial examples. Techniques such as data augmentation, where training data can be expanded by creating modified versions of existing data or synthesizing new data, have become commonplace. For instance, in 2023 IEEE competition, Kreuzer’s team employed methods like additive white Gaussian noise, circular shift, and random amplitude scaling to increase the volume of training data~\cite{kreuzer20231}. Moreover, regularization techniques like L1 and L2 regularization have been instrumental in preventing overfitting and enhancing stability, making the model less sensitive to small fluctuations in input data. What’s more, ensemble methods have been increasingly recognized for the contribution to model robustness in PHM. A comprehensive analysis of recent studies shows that out of 59 research papers, nine utilized various ensemble techniques - including tree-based methods, ensemble LSTM~\cite{aydemir2021ensemble}, soft voting ensemble~\cite{kim2023domain}, transfer ensemble learning~\cite{tian2023novel}, ensemble regression~\cite{rao2020structure}, and ensemble CNN-based~\cite{lee2023ensemble} approaches - across six different PHM data challenge competitions. Additionally, adversarial training, which involves training models on both regular and adversarial data, has been recognized for its potential to fortify models against adversarial attacks~\cite{wu2020ss,kim2023domain,qiu2023deep}.
Moreover, it is also important to consider probabilistic machine learning techniques, such as Bayesian networks, Gaussian processes, and probabilistic graphical models, which incorporate probability theory into the modeling process to handle uncertainty issues. They are crucial for dealing with uncertainties and improving the robustness of PHM models~\cite{ghahramani2015probabilistic,murphy2022probabilistic,hazra2024reliability}.

\begin{figure*}[h!]
\centering\includegraphics[width=0.65\linewidth]{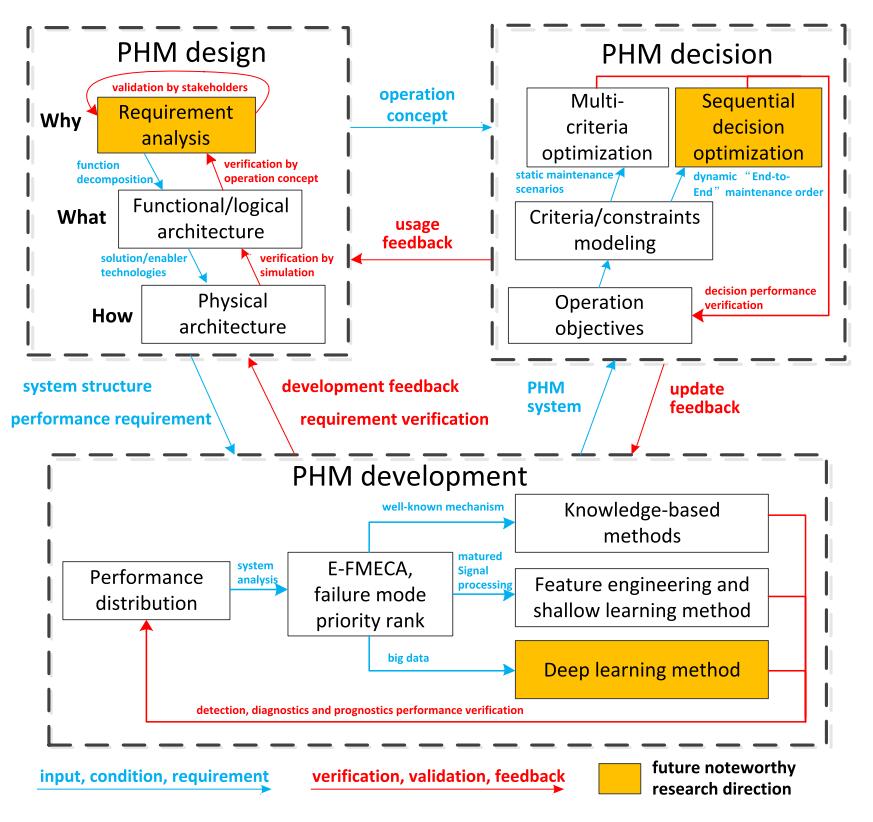}
\caption{Interactions of PHM Phases of Design, Development, and Decision ($DE^3$) \protect\cite{hu2022prognostics}}
\label{fig:5}
\end{figure*}

Alongside robustness, model generalization refers to developing models that perform reliably on new, unseen data. One of the techniques for improving generalization is cross-validation, which involves dividing the training data into several subsets and each time using one of the subsets as the validation data and others as the training data for model validation during the training process. In 2021 PHM NA, researchers opted for k-fold repeated random subsampling validation to address its limitation, wherein the size of the validation set diminishes as the number of folds (k) increases~\cite{solis2021stacked}. In addition to cross-validation, transfer learning, and domain adaptation methods have been crucial in maintaining model effectiveness and generalization when data from the target domain differs from the source domain. The specific implementations in PHM competitions have been detailed in the earlier subsubsection titled “Domain Shift”. Outside of PHM data challenges, some novel approaches are proposed to deal with model generalization issues. Matthew Russell and Peng Wang adopted a domain adversarial transfer learning method inspired by generative adversarial networks, utilizing a 1D CNN architecture to predict tool wear on unseen domains using NASA's milling dataset~\cite{wang2020domain}. Ding et al. developed a multi-source domain generalization learning approach (GRU + Transformer) that can effectively learn useful degradation feature representations from various run-to-failure datasets of internal combustion engine journal bearings across different conditions and predict unseen working conditions well~\cite{ding2023multi}. Furthermore, Ding et al. proposed an adversarial out-domain augmentation (AOA) framework for predicting the RUL of bearings under unseen conditions. The effectiveness of this AOA-based RUL prediction was validated using IEEE PHM Challenge 2012 and XJTU-SY run-to-failure datasets, illustrating its robustness in domain generalization for predictive maintenance~\cite{ding2023domain}.

\subsection{Limitations of Current PHM Competitions and Opportunities}

\subsubsection{A Lack of Multi-Modal Machine Learning Analysis}
Our review of PHM data challenge competitions reveals a reliance on single-modality data in the competitions, such as pressure signals, currents, vibrations, or images, without incorporating multi-modal datasets for fault diagnosis and prognosis. Multi-modal machine learning (MMML) in PHM refers to capturing complementary information from multiple data sources (different types) to achieve a more comprehensive and precise evaluation of PHM tasks~\cite{ramachandram2017deep,tsanousa2022review}. Despite its potential, it is still underexplored in the PHM domain. 

Recently, Jiang et al. leveraged two modality data (vibration and current signals) to develop deep belief networks (DBNs) for diagnosing wind turbine gearbox faults~\cite{jiang2019intelligent}. Su et al. proposed an MMML model using parametric specifications, text descriptions, and images of vehicles to predict five vehicle rating scores~\cite{su2023multi}. Additionally, Fan et al. evaluate several MMML strategies to create a comprehensive PHM system for coolant pumps in commercial heavy-duty vehicles, utilizing data from onboard signals, multi-dimensional histograms, and categorical variables~\cite{fan2023evaluation}. Wang et al. proposed a novel method for feature fusion in multimodal data (vibration and torque signals), applying it to diagnose the bearings faults~\cite{wang2021novel}. For future PHM data challenges, the provision of open-source multi-modal datasets would empower researchers to investigate and apply more advanced MMML techniques, potentially leading to advancements in the context of PHM.

\subsubsection{A Lack of Adopting Systematic Approaches for Effective PHM Systems Construction}

Our analysis of ML and DL methods across nine open-source industrial datasets has revealed the advantages of ML methods applied in PHM. Nevertheless, the nature of the competition tends to prioritize solutions that chiefly enhance accuracy, potentially at the expense of a systematic approach, reusability, and methodological inheritance. It is therefore vital to pursue systematic methodologies for constructing effective PHM systems that go beyond the competitive framework. This should involve thorough research and analysis of open-source datasets to advance ML and DL strategies, aiming not just for competition success but also for benchmarking and comparative analysis.

Souza et al. devised an ML-based, data-oriented pipeline for constructing a Prognosis and Health Management System (PHMS) focused on RUL prediction, utilizing semi-supervised ML with Autoencoder, XGBoost, and SHAP method~\cite{souza2023machine}. As shown in Figure~\ref{fig:5}, Hu et al. offered a new perspective on reviewing PHM efforts by proposing a division of the PHM lifecycle into DEsign, DEvelopment, and DEcision ($DE^3$) phases, and showcasing the important activities and challenges within these stages~\cite{hu2022prognostics}. Additionally, Lee et al. introduced a novel SoQ methodology for multi-stage manufacturing processes. It can help to analyze multi-parameter influences on product quality and model inter-process relationships in multi-stage manufacturing systems~\cite{lee2022stream}. Moving forward, developing novel, systematic approaches for PHM systems should be encouraged in future PHM data challenge competitions. These efforts can augment the systematization and applicability of ML and DL approaches, thereby expanding their utility beyond the confines of the competition-centric paradigm. Such advancements promise to narrow the divide between academic research and industrial practice, facilitating the broader adoption of data-driven ML across various industrial contexts.

\begin{figure*}[h!]
\centering\includegraphics[width=\linewidth]{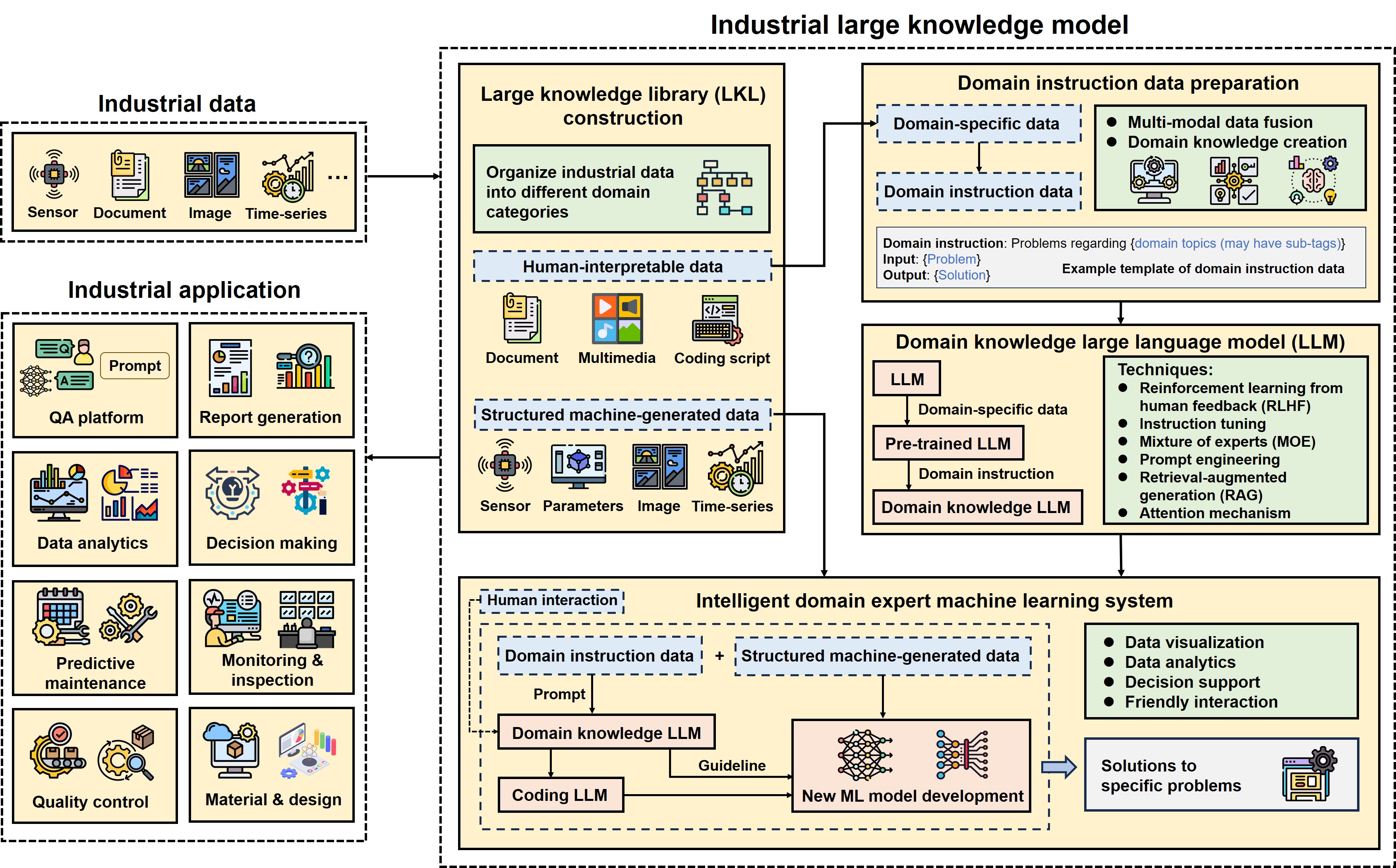}
\caption{Industrial Large Knowledge Model Framework \protect\cite{lee2024unified}}
\label{fig:6}
\end{figure*}

\section{Prospects}
There are still some research directions that deserve deeper investigation and exploration by the research community going forward.

\textbf{(1) A Need for Open-Source Multi-Modal Datasets.}
In the PHM domain, the availability of multi-modal datasets is notably limited. While prior research has leveraged diverse modal information, including vibration, current, or torque signals for diagnosing issues in wind turbine gearboxes, bearings, or other industrial products~\cite{jiang2019intelligent,wang2021novel,fan2023evaluation}, these datasets are all private. This restriction to some extent hampers the capacity for broad-based development of MMML approaches. To overcome this challenge, there is a need for the PHM community to collaboratively establish and maintain multi-modal industrial datasets, enriched with high-quality data and labels. This would involve the collection, alignment, and annotation of multi-modal data with PHM-centric attributes. Moreover, providing latent representations or pre-trained embeddings, if possible, can also accelerate and efficiently train new MMML models and facilitate knowledge transfer across various PHM tasks, ultimately benefiting the whole PHM community.

\textbf{(2) Development of Multi-Modal Machine Learning Approaches.}
Furthermore, the investigation of a broader range of MMML techniques is highly encouraged. On the one hand, for time series data of a single modality, it may be feasible to extract features representative of different modalities from the time series data itself, such as the time domain (origin signal), frequency domain (FFT), PSD, STFT, etc. This approach could lead to the preliminary training of unimodal ML models on each single modality, followed by the exploration of MMML strategies. On the other hand, MMML methodologies necessitate more effective representation learning and information alignment techniques—the former concerning the efficient encoding of single modality data, and the latter focusing on the enhanced analysis and fusion of multimodal information for effective PHM prediction or classification tasks. Although simple concatenation is a common method for data fusion in MMML, emerging fusion techniques, such as attention based or transformer based mechanisms~\cite{vaswani2017attention}, deserve further exploration. These advanced methods have the potential to effectively capture implicit feature alignments across modalities and facilitate cross-modal synthesis~\cite{mansimov2015generating,xu2018attngan}. Yet, research on MMML within the PHM field remains underexplored, highlighting a significant opportunity to explore.

\textbf{(3) Further Exploration in Machine Learning Model Interpretability.}
The adoption of DL in PHM has underscored the need for models that are not only high-performing but also interpretable. Techniques such as advanced data visualization and XAI methods are emerging as key tools in explaining the outputs of ML models~\cite{linardatos2020explainable,arrieta2020explainable}. These methods are anticipated to provide industries and academia with clearer insights into the decision-making processes of PHM models, thereby building trust and facilitating more informed decision-making~\cite{solis2023soundness}. However, current XAI methods predominantly address the interpretability of models using tabular data, text, and images as inputs, leaving a gap in methods tailored for time series data. Moreover, most of the current interpretability methods are applied to unimodal ML models, and the interpretability of MMML models has not been explored. Addressing these gaps can help to balance the performance and interpretability of ML models. 

\textbf{(4) Novel Transfer Learning and Domain Adaptation Techniques Development for Model Robustness, and Generalization.}
Alongside interpretability, the robustness and generalization of ML models are also important. Novel approaches in transfer learning and domain adaptation can be further developed to ensure models are resilient to data variability and operational uncertainties and capable of adapting to new, unseen scenarios~\cite{azari2023systematic}. Currently, research on transfer learning in the PHM domain predominantly addresses fault diagnosis, with only a few studies exploring prognosis. Looking forward, how to utilize the power of transfer learning for prediction problems is still a critical issue. Additionally, cross-modal transfer learning (CMTL) emerges as a critical area of interest in PHM, aiming to improve the knowledge transfer between distinct domains. Moreover, the challenge of collecting a sufficiently large, labeled dataset is a significant barrier in practical applications. The development of unsupervised and semi-supervised transfer learning techniques may help to address this issue. 

\textbf{(5) Potential Utilization of Large Language Models (LLMs) and Industrial Large Knowledge Models (ILKMs).}
Recent advancements in large language model technologies have shown remarkable abilities in natural language processing and related tasks, hinting at the potential for general artificial intelligence applications~\cite{zhao2023survey}. Leveraging these cutting-edge technologies could lead to new changes in PHM domain. Yang et al. introduced a novel benchmark dataset focused on Question Answering (QA) in the industrial domain and proposed a new model interaction paradigm, aimed at enhancing the performance of LLMs in domain-specific QA tasks~\cite{wang2023empower}. This approach signifies a substantial stride in customizing LLMs for more specialized, industry-oriented applications. Meanwhile, Li's team systematically reviewed the current progress and key components of ChatGPT-like large-scale foundation (LSF) models, and provided a comprehensive guide on adapting these models to meet the specific needs of PHM, underscoring the challenges and opportunities for future development~\cite{li2023chatgpt}. Moreover, as shown in Figure~\ref{fig:6}, Lee's team proposed an Industrial Large Knowledge Model (ILKM) framework that aims to solve complex challenges in intelligent manufacturing by combining LLMs and domain-specific knowledge~\cite{lee2024unified}. Therefore, integrating specialized domain knowledge with LLM technology presents a good opportunity to develop more effective ML models, potentially leading to better solutions for challenges in PHM.

\section{Conclusion}

In summary, ML gradually becomes a cornerstone in PHM, reflecting the potential for innovative advancements in future PHM development. This paper serves as a valuable resource for both academic and industry professionals in the PHM domain, offering a unified ML framework in PHM and a comprehensive overview of the current state-of-the-art ML approaches for diagnostics and prognostics of industrial systems using industrial open-source data from recent PHM data challenge. Based on two primary research task categories: "Detection \& Diagnosis" and "Assessment \& Prognosis", we provide a detailed explanation of the problems, tasks, challenges, and relevant ML approaches to each competition. Furthermore, we summarize common challenges, including data-related and model-related issues, and analyze the solutions to address these challenges. Moreover, we evaluate the limitations of these PHM data challenge competitions and suggest future directions that PHM data challenge competition could focus on. Finally, we prospect five potential research directions in the application of data-driven ML within PHM, encompassing
a need for open-source multi-modal datasets, development of MMML approaches, further exploration of ML model interpretability, improving the robustness, and generalization of ML models, and utilization the potential of LLMs and ILKMs.

\section*{Nomenclature}
\begin{tabular}{ l  l }
	$AI$			&Artificial Intelligence\\ 
	$Bi-LSTM$			&Bidirectional Long Short-Term Memory\\ 
	$CNN$		&Convolutional Neural Network\\ 
	$DANN$			&Domain Adversarial Neural Networks\\  
	$DL$			&Deep Learning\\ 
	$DTW$			&Dynamic Time Warping\\ 
	$FCM$			&Fuzzy C-Means\\ 
	$FCN$  	   		&Fully Convolutional Network\\ 
	$FIR$	   		&Feature Importance Ranking\\  
	$GAN$   	   	&Generative Adversarial Networks\\ 
	$GRU$      	&Gated Recurrent Units\\
        $ILKM$      	&Industrial Large Knowledge Model\\ 
        $KNN$      	&K-Nearest Neighbors\\ 
        $LDA$      	&Linear Discriminant Analysis\\ 
        $LightGBM$      	&Light Gradient Boosting Machine\\ 
        $LIME$      	&Local Interpretable Model-Agnostic Explanations\\
        $LKL$      	&Large Knowledge Library\\ 
        $LLM$      	&Large Language Model\\ 
        $LSTM$      	&Long Short-Term Memory\\ 
        $ML$      	&Machine Learning\\ 
        $MLP$      	&Multi-Layer Perceptron\\ 
        $MMD$      	&Maximum Mean Discrepancy\\ 
        $PCA$      	&Principal Component Analysis\\ 
        $PDF$      	&Probability Density Function\\ 
        $PHM$      	&Prognostics and Health Management\\ 
        $PLS$      	&Partial Least Squares\\ 
        $PSD$      	&Power Spectral Density\\ 
        $QA$      	&Question Answering\\ 
        $RNN$      	&Recurrent Neural Network\\
        $RUL$      	&Remaining Useful Life\\
        $SHAP$      	&SHapley Additive exPlanations\\
        $SMOTE$      	&Synthetic Minority Oversampling TEchnique\\
        $SoQ$      	&Stream-of-Quality\\
        $STFT$      	&Short-Time Fourier Transform\\
        $SVM$      	&Support Vector Machine\\
        $SVR$      	&Support Vector Regression\\
        $TCN$      	&Temporal Convolutional Network\\
        $XAI$      	&Explainable Artificial Intelligence\\
        $XGBoost$      	&Extreme Gradient Boosting\\
 \end{tabular}

\bibliographystyle{abbrv}
\bibliography{computers_in_industry}

\section*{Appendix}
\begin{table*}[!htbp]
\LARGE
\caption{Task Description, References, and Website Links for PHM Data Challenge Competitions}
\label{table:8}
\begin{center}
\resizebox{18cm}{!}{
\renewcommand{\arraystretch}{2.0}
\begin{tabular}{llll}
\toprule
\multicolumn{1}{c}{\textbf{PHM Data Challenge Competition}} &
  \multicolumn{1}{c}{\textbf{Task Description}} &
  \multicolumn{1}{c}{\textbf{Reference}} &
  \multicolumn{1}{c}{\textbf{URL}} \\ \hline
\begin{tabular}[c]{@{}l@{}}2018 PHM NA\\ Ion Mill Etching System\end{tabular} &
  \begin{tabular}[c]{@{}l@{}}(1) Diagnose failures\\ (2) Determine time remaining until next failure\end{tabular} & 
  \begin{tabular}[c]{@{}l@{}}\cite{vishnu2018recurrent,singh2018concurrent,huang2018remaining}\\ \cite{he2019failure,zheng2021cross,wu2021remaining,kim2021challenges,liu2021two}\\
  \cite{hsu2022temporal,zhao2022deep,lorenti2023predictive,cicak2023handling}\\
  \end{tabular} &
  \begin{tabular}[c]{@{}l@{}}\href{https://phmsociety.org/conference/annual-conference-of-the-phm-society/annual-conference-of-the-prognostics-and-health-management-society-2018-b/phm-data-challenge-6/}{https://phmsociety.org/conference/annual-conference-of-the-phm-soc}\\
  \href{https://phmsociety.org/conference/annual-conference-of-the-phm-society/annual-conference-of-the-prognostics-and-health-management-society-2018-b/phm-data-challenge-6/}
  {iety/annual-conference-of-the-prognostics-and-health-management}\\
  \href{https://phmsociety.org/conference/annual-conference-of-the-phm-society/annual-conference-of-the-prognostics-and-health-management-society-2018-b/phm-data-challenge-6/}
  {-society-2018-b/phm-data-challenge-6/}\end{tabular} \\ 
\begin{tabular}[c]{@{}l@{}}2019 PHM NA\\ Fatigue Crack\end{tabular} &
  (1) Estimate crack length &
  \begin{tabular}[c]{@{}l@{}}\cite{karimian2020neural,kong2020hybrid}\\ \cite{youn2020fatigue,rao2020structure}\\
  \end{tabular} &
  \href{https://data.phmsociety.org/2019datachallenge/}{https://data.phmsociety.org/2019datachallenge/} \\
\begin{tabular}[c]{@{}l@{}}2020 PHM EU\\ Filtration System\end{tabular} &
  (1) Predict RUL of the filtration system &
  \begin{tabular}[c]{@{}l@{}}\cite{lomowski2020method,beirami2020data,ince2020remaining,vu2021deep}\\
  \cite{lee2021data,falcon2022neural,tian2023novel,ince2023joint}\\ 
  \end{tabular} &
  \href{http://phmeurope.org/2020/data-challenge-2020}{http://phmeurope.org/2020/data-challenge-2020} \\
\begin{tabular}[c]{@{}l@{}}2021 PHM EU\\ Manufacturing Production Line\end{tabular} &
  \begin{tabular}[c]{@{}l@{}}(1) Fault detection\\ (2) Fault diagnosis (Classification)\\ (3) Root cause identification\end{tabular} &
  \begin{tabular}[c]{@{}l@{}}\cite{de2021divide,ince2021fault,aimiyekagbon2021rule}\\ \cite{aydemir2021ensemble,ramezani2021explainable,alfarizi2022extreme,tian2022high}\\
  \end{tabular} &
  \href{https://github.com/PHME-Datachallenge/Data-Challenge-2021}{https://github.com/PHME-Datachallenge/Data-Challenge-2021} \\
\begin{tabular}[c]{@{}l@{}}2021 PHM NA\\ Turbofan Engine\end{tabular} &
  (1) RUL prediction in a fleet of aircraft engines &
  \begin{tabular}[c]{@{}l@{}}\cite{lovberg2021remaining,solis2021stacked,devol2021inception,devol2022evaluating,solis2023soundness}\\
  \cite{cohen2023fault,cohen2023trust,cohen2023shapley}\\ 
  \end{tabular} &
  \href{https://data.phmsociety.org/2021-phm-conference-data-challenge/}{https://data.phmsociety.org/2021-phm-conference-data-challenge/} \\
\begin{tabular}[c]{@{}l@{}}2022 PHM EU\\ Printed Circuit Boards\end{tabular} &
  \begin{tabular}[c]{@{}l@{}}(1) Prediction of defects in AOI based on SPI\\ (2) Prediction of human inspection label\\ (3) Prediction of human-assigned repair label\end{tabular} &
  \begin{tabular}[c]{@{}l@{}}\cite{gaffet2022hierarchical,taco2022novel,tang2022prediction}\\
  \cite{schmidt2022application,lee2022stream,mirzaei2023application}\\ 
  \end{tabular} &
  \href{https://github.com/PHME-Datachallenge/Data-Challenge-2022}{https://github.com/PHME-Datachallenge/Data-Challenge-2022} \\
\begin{tabular}[c]{@{}l@{}}2022 PHM NA\\ Rock Drill\end{tabular} &
  (1) Multiclass fault classification for rock drills &
  \begin{tabular}[c]{@{}l@{}}\cite{oh2023hybrid,kim2023domain,minami2023novel}\\
  \cite{ling2023hydraulic,taco2023novel}\\ 
  \end{tabular} &
  \href{https://data.phmsociety.org/2022-phm-conference-data-challenge/}{https://data.phmsociety.org/2022-phm-conference-data-challenge/} \\
\begin{tabular}[c]{@{}l@{}}2023 PHM AP\\ Spacecraft Propulsion System\end{tabular} &
  \begin{tabular}[c]{@{}l@{}}(1) Diagnose normal, bubble anomalies \\ (2) Diagnose solenoid valve faults\\ (3) Diagnose unknown abnormal cases\end{tabular} &
  \begin{tabular}[c]{@{}l@{}}\cite{minami2023phm,berndt1994using,lee2023hybrid}\\
  \cite{kato2023anomaly,aimiyekagbon2023expert}\\ 
  \end{tabular} &
  \href{https://phmap.jp/program-data/}{https://phmap.jp/program-data/} \\
\begin{tabular}[c]{@{}l@{}}2023 IEEE\\ Gearbox\end{tabular} &
  (1) Fault diagnosis (Classification) &
  \begin{tabular}[c]{@{}l@{}}\cite{lee2023ensemble,shen2023gear}\\
  \cite{kreuzer20231,sobha2023comprehensive}\\ 
  \end{tabular} &
  \begin{tabular}[c]{@{}l@{}}\href{http://www.ieeereliability.com/PHM2023/index.html}{http://www.ieeereliability.com/PHM2023/index.html} \\ or \href{http://www.ieeereliability.com/PHM2023/assets/data-challenge.pdf}{http://www.ieeereliability.com/PHM2023/assets/data-challenge.pdf}\end{tabular} \\ \bottomrule
\end{tabular}}
\end{center}
\end{table*}

\end{document}